\documentclass{article}

\usepackage{arxiv}

\usepackage{algorithm}
\usepackage{algpseudocode}
\def\bSig\mathbf{\Sigma}

\usepackage{amsmath,amssymb}
\DeclareMathOperator{\E}{\mathbb{E}}
\usepackage{natbib}
\newcommand{\indep}{\perp \!\!\! \perp}
\usepackage{bbm}
\usepackage{enumitem}
\usepackage{graphicx}
\usepackage{rotating}
\usepackage{url}
\usepackage{verbatim}
\usepackage{hyperref}
\usepackage{rotating} 
\usepackage{caption}

\title{Causal machine learning for heterogeneous treatment effects in the presence of missing outcome data}

\author{Matthew Pryce$^{1,*}$, Karla Diaz-Ordaz$^{2}$, Ruth H. Keogh$^{1}$, Stijn Vansteelandt$^{3}$ \vspace{0.5cm}\\
$^{1}$Department of Medical Statistics, London School of Hygiene \& Tropical Medicine, London, WC1E 7HT,\\ United Kingdom \\
$^{2}$Department of Statistical Science, University College London, London, United Kingdom\\
$^{3}$Department of Applied Mathematics, Computer Science, and Statistics, 
Ghent University, \\Ghent, Belgium \vspace{0.5cm}\\
matthew.pryce@lshtm.ac.uk}

\begin{document}
\maketitle
\begin{abstract}
When estimating heterogeneous treatment effects, missing outcome data can complicate treatment effect estimation, causing certain subgroups of the population to be poorly represented. In this work, we discuss this commonly overlooked problem and consider the impact that missing at random (MAR) outcome data has on causal machine learning estimators for the conditional average treatment effect (CATE). We propose two de-biased machine learning estimators for the CATE, the mDR-learner and mEP-learner, which address the issue of under-representation by integrating inverse probability of censoring weights into the DR-learner and EP-learner respectively. We show that under reasonable conditions, these estimators are oracle efficient, and illustrate their favorable performance through simulated data settings, comparing them to existing CATE estimators, including comparison to estimators which use common missing data techniques. We present an example of their application using the GBSG2 trial, exploring treatment effect heterogeneity when comparing hormonal therapies to non-hormonal therapies among breast cancer patients post surgery, and offer guidance on the decisions a practitioner must make when implementing these estimators.
\end{abstract}



\section{Introduction}
\label{s:intro}

When evaluating the effect of an intervention, investigators are often interested in how the effect may vary within a target population. One approach used to explore treatment effect heterogeneity for a binary intervention is to estimate the conditional average treatment effect (CATE), defined as $\theta(x) = \E[Y(1)|X=x] - \E[Y(0)|X=x]$, where $Y(0)$ and $Y(1)$ are potential outcomes under the two levels of the treatment \citep{RN16} and $X$ represents the individual (pre-treatment) characteristics in which heterogeneity is of interest.

In recent years, many estimators for the CATE have been proposed, with the focus turning towards non-parametric estimators, using machine learning (ML) to estimate complex functions of high dimensional data \citep{RN8,RN10,RN9,RN50}. Of these estimators, each require that the training data be fully-observed and no data be missing. In this paper, we relax this requirement and propose two novel CATE estimators, the mDR-learner and mEP-learner, which demonstrate how causal ML estimators can be constructed when outcome data is missing at random (MAR). 

MAR outcome data occurs frequently in practice, typically arising when individuals are lost to follow up. When it occurs, the observed data may no longer represent the target population, and subgroups that have high levels of drop-out can be under-represented. This presents a challenge for existing non-parametric CATE estimators, which do not address this under-representation and are prone to producing biased estimates of the CATE within these under-represented subgroups. To overcome this, some authors propose using CATE estimators in combination with established missing data techniques, such as imputing missing outcomes \citep{RN19,RN20}, or re-weighting the population using inverse probability of censoring weights (IPCW) \citep{RN1,RN26}. However, when implementing these approaches using non-parametric, machine learning techniques, the inherent slow convergence of ML algorithms can introduce errors into the IPCW/imputation predictions, which then propagate through to the CATE estimates. 

Our work aims to overcome this issue, with the mDR-learner and mEP-learner robustly incorporating IPCWs into the DR-learner \citep{RN10} and EP-learner \citep{RN50} respectively. We show that these estimators are oracle efficient under reasonable conditions and demonstrate their empirical performance through a simulation study (Section \ref{Sims}). We then illustrate their application, exploring treatment effect heterogeneity within the GBSG2 trial (Section \ref{GBSG2}) and finish by discussing potential extensions (Section \ref{Discussion}).

\section{Background} \label{Background}
\subsection{Setting}\label{setting}

We define a sample of $n$ independent and identically distributed (i.i.d.) observations drawn from $O = (Z,A,CY)$, where $A$ is a binary treatment, $A \in \{0,1\}$ and $Y$ is a continuous outcome. In this setting, we allow outcome data to be missing and define $C$ as an indicator of outcome missingness, $C \in \{0,1\}$, with $C=1$ indicating that an individual's outcome is non-missing. We then define $Z$ to be a set of baseline covariates which contains all confounders between the treatment $A$ and the outcome $Y$, and also contains sufficient information for missing outcomes to be assumed MAR (i.e. after controlling for $Z$ and $A$, the missingness indicator $C$ and outcome $Y$ are independent). Finally, we focus on learning the CATE, $\theta(x) = \E[Y(1)|X=x] - \E[Y(0)|X=x]$, which is conditional on $X\subseteq Z$, where $X$ are the covariates for which heterogeneity is of interest. 

\subsection{Estimating the CATE using ML - No missing outcome data} \label{Background_nomiss}

We first consider how causal ML estimators are constructed when no outcome data is missing. Under the standard causal assumptions of (A1) consistency, (A2) no unmeasured confounding, $Y(a)\indep A|Z$ for $a\in\{0,1\}$ and (A3) positivity of treatment exposure, $0<P(A=1|Z)<1$ (with probability 1) \citep{RN57}, the CATE can be identified as: 
\begin{equation} \label{Ident_CATE_nomiss_out}
  \theta(x) = \E[\E[Y|A=1,Z] - \E[Y|A=0,Z]|X=x],  
\end{equation}
\begin{center}
    or
\end{center}
\begin{equation} \label{Ident_CATE_nomiss_prop}
  \theta(x) = \E\left[\left.\frac{AY}{P[A=1|Z]}-\frac{(1-A)Y}{1-P[A=1|Z]}\right|X=x\right].  
\end{equation}

ML can be used to estimate the conditional expectations/probabilities (nuisance functions) found in eq. (\ref{Ident_CATE_nomiss_out}) or eq. (\ref{Ident_CATE_nomiss_prop}), with these estimators commonly referred to as plug-in estimators. One common example of a plug-in estimator is the T-learner \citep{RN8}, which requires $X=Z$ and estimates the conditional expectations from eq. (\ref{Ident_CATE_nomiss_out}) in the subsets of individuals who are treated/untreated, $\mu^1(Z)=\E[Y|A=1,Z]$ and $\mu^0(Z)=\E[Y|A=0,Z]$, taking their difference to obtain CATE estimates. Alternatively, plug-in estimators based on eq. (\ref{Ident_CATE_nomiss_prop}) are often referred to as inverse probability of treatment weight (IPTW) estimators, as they require the estimation of the propensity score, $\pi(Z)=P[A=1|Z]$ \citep{RN10}. 

While such ML-based plug-in estimators are simple to understand and implement, they are also prone to bias. This arises as ML algorithms use regularization to avoid over-fitting to the training data, reducing the rate at which these estimators converge towards the true parameter. This introduces non-negligible bias when fit using finite samples, with the errors in these nuisance function estimates (e.g. $\hat{\mu}^1(Z), \hat{\mu}^0(Z)$ or $\hat{\pi}(Z)$) directly propagating through to the estimates of the CATE, known as ``plug-in bias'' \citep{RN60}. Furthermore, the T-learner does not ensure an optimal bias-variance trade-off is made for the CATE, as it optimizes predictions for their outcome functions rather than for the CATE itself \citep{RN10}. See Appendix \ref{ill_ex_CC} for an illustrative example. 

Estimators that target the CATE directly and that are less sensitive to nuisance model estimation avoid these issues. Estimators of this form are typically constructed using the efficient influence function (EIF) of the estimand of interest, where the EIF represents how sensitive a measure of prediction error is to changes in the data generating distribution. The EIF offers a useful tool for constructing estimators, allowing estimators to be designed which are approximately insensitive to small changes in its nuisance functions, meaning estimation errors in the outcome functions or propensity score affect errors in the estimates of the target function only through their product. Unfortunately, the EIF of the CATE is generally not well defined (i.e., has infinite variance) whenever it depends on continuous variables \citep{RN29}. Instead, estimators which achieve these properties can be constructed based on the EIF of a well-chosen loss function, in this case, a measure of counterfactual prediction error \citep{RN22,RN60}. Two estimators constructed in this way are the DR-learner \citep{RN10} and EP-learner \citep{RN50}.

The DR-learner is a model-agnostic CATE estimator, meaning the user can choose any estimation strategy, including data adaptive methods, when estimating any functions within it. It is derived using a two-step procedure \citep{RN10}, with the first step calculating pseudo-outcomes, $Y_{DR}$, using the EIF of the mean square error (MSE) for the CATE:
\begin{equation} \label{ATE_EIF_no_MISS}
    Y_{DR} = \frac{\left(A-\pi(Z)\right)}{\pi(Z)\left(1-\pi(Z)\right)}\left\{Y - \mu^{A}(Z)\right\} + \mu^1(Z) - \mu^0(Z),
\end{equation}
where $\hat{\mu}^{A}(Z) = A\cdot\hat{\mu}^{1}(Z) + (1-A)\cdot\hat{\mu}^{0}(Z)$. The second step then learns the CATE by regressing the pseudo-outcomes on the covariate set in which heterogeneity is of interest, $X$. See Appendix \ref{App_DR} for the full algorithm. By using a pseudo-outcome regression, the DR-learner targets the CATE directly, and can benefit from faster convergence rates when the CATE is smooth. Additionally, these pseudo-outcomes are derived by considering the components of the mean squared error (MSE) of the CATE that depend on $\theta(X)$, $\psi_{CATE} = \E\left[\theta(X)^2 - 2\theta(X)(\mu^1(Z)-\mu^0(Z))\right]$, using the EIF of this risk function, 
\begin{align} \label{MSE_IF_nomiss}
    \phi = \left(\theta^2(X) - \theta(X)\left(\mu^1(Z)-\mu^0(Z)\right)\right) + \frac{2\theta(X)(A-\pi(Z))}{\pi(Z)(1-\pi(Z))}\{Y-\mu^A(Z)\} - \psi_{CATE},
\end{align}
to define its pseudo-outcomes such that the sample average of the drift term (second term in eq. (\ref{MSE_IF_nomiss})) goes to 0. These pseudo-outcomes closely resemble the EIF from the average treatment effect (ATE) and consist of the plug-in estimates and a weighted error term. By defining the pseudo-outcomes in this way, the DR-learner ensures that the gradient of the MSE risk function, with respect to the CATE, is less sensitive to errors in its nuisance functions. It also allows the DR-learner to achieve oracle efficiency, meaning that when the product of the convergence rates for the outcome predictions and propensity score estimates is faster than the rate of the oracle learner, it performs asymptotically as if the nuisance functions were known \citep{RN10}. However, the convergence rate of the DR-learner will still depend on the convergence rate of its pseudo-outcome regression, which will vary itself depending on the complexity of the CATE.  

Despite its desirable properties, the DR-learner can be sensitive to extreme propensity scores, with the IPTWs used in its pseudo-outcomes causing the pseudo-outcomes to grow infinitely large when propensity score estimates are near 0/1. To prevent this, some trim the propensity score estimates, therefore stopping the pseudo-outcomes from growing too large. However, this introduces bias to the propensity score estimates which can propagate through to the CATE estimates themselves. For this reason, we discuss an alternative approach known as infinite-dimensional targeting (iTMLE) \citep{RN68,RN55,RN50}. This technique has been developed for counterfactual outcome prediction \citep{RN55}, and tends to have less sensitivity to extreme propensity scores as its targeted learning framework can moderate the impact of outlying propensity score estimates. In this paper we discuss a very similar approach known as the EP-learner (efficient plug-in learner) \citep{RN50}, which uses iTMLE in the context of causal contrasts, including CATE estimation. 

The EP-learner is a model-agnostic estimator which also uses a two step procedure, first deriving pseudo-outcomes which are later regressed on $X$. However, the EP-learner uses an iTMLE procedure to generate its pseudo-outcomes. This iTMLE procedure plays a similar role to the one-step correction used in the DR-learner, and is motivated by the EIF of the MSE risk function for the CATE, eq. (\ref{MSE_IF_nomiss}). Using this EIF, the iTMLE procedure aims to update the initial plug-in outcome estimates, $\hat{\mu}^0(Z)$ and $\hat{\mu}^1(Z)$ using targeted learning such that the sample average of the drift term (second term) in eq. (\ref{MSE_IF_nomiss}), goes to 0. Yet, since the drift term in eq. (\ref{MSE_IF_nomiss}) contains an infinite dimensional $\theta(X)$ (when any variables in $X$ are continuous), the classical targeted learning procedure which regresses outcomes against a scalar (known as a clever covariate) would fail to set the sample average of the drift term to 0. To resolve this, the iTMLE procedure defines a vector of univariate basis functions, referred to as a sieve, $\varphi(X)$, which it uses within the targeting step, approximating $\theta(X)$ and allowing the updates to the plug-in estimates to vary by $X$. 

The targeting step then works by regressing the outcomes $Y$ on the sieve basis, $\varphi(X)$, in a weighted linear regression, with an offset $\mu^A(Z)$, and weight, $\hat{H}(A,Z)=\frac{A}{\hat{\pi}(Z)} + \frac{1-A}{\left(1-\hat{\pi}(Z)\right)}$, suggested by the EIF of the risk function, eq. (\ref{MSE_IF_nomiss}). After fitting this model, efficient plug-in estimates, $\hat{\mu}^{0*}(Z)$ and $\hat{\mu}^{1*}(Z)$ are obtained by adding/subtracting the estimated linear predictor from this model to the plug-in outcome estimates (See Appendix \ref{App_EP} for the full algorithm). As the updated outcome predictions are defined such that the sample average of the drift term in eq. (\ref{MSE_IF_nomiss}) converges to 0, the EP-learner achieves the same oracle efficiency properties as the DR-learner. Additionally, by using iTMLE, extreme pseudo-outcomes are less common and the CATE estimates should be more stable \citep{RN50}. 

In addition, both learners require the data which is used in the second stage optimization is i.i.d. To achieve this, sample splitting is used when estimating the nuisance functions, preventing correlations from being introduced into the pseudo-outcomes. In this paper, we implement a K-fold cross-fitting procedure found in Appendix \ref{App_DR} and \ref{App_EP}.  


\subsection{Estimating the CATE using ML - With missing outcome data} \label{Background_miss}

Now we consider how MAR outcome data impacts the existing estimators. To identify the CATE when outcome data is MAR, we require assumptions A1-A3 from Section \ref{Background_nomiss} and two additional assumptions on the missingness mechanism: (A4) Outcomes are MAR, $Y\indep C|A, Z$; (A5) Positivity of outcomes being non-missing, $0<P(C=1|A,Z)$, with probability 1. Under assumptions A1-A5, the CATE can be identified as eq. (\ref{ident_CATE_miss_out}) and eq. (\ref{Ident_CATE_miss_prop}).
\begin{equation} \label{ident_CATE_miss_out}
   \theta(x) = \E[\E[Y|A=1,C=1,Z] - \E[Y|A=0,C=1,Z]|X=x] ,
\end{equation}
\begin{center} or
\end{center}
\begin{equation} \label{Ident_CATE_miss_prop}
  \theta(x) = \E\left[\left.\frac{CAY}{P[C=1|A,Z]P[A=1|Z]}-\frac{C(1-A)Y}{P[C=1|A,Z](1-P[A=1|Z])}\right|X=x\right].  
\end{equation}

The no unmeasured confounding assumption (A2) and MAR assumption (A4) need not be conditional on the same covariates, however, for simplicity, we define both assumptions to be conditional on the same set of covariates, $Z$. Using equations (\ref{ident_CATE_miss_out}) and (\ref{Ident_CATE_miss_prop}) we see that when all of the covariates in $Z$ are present, the CATE can be estimated using the observed data. Consequently, the existing CATE estimators, such as the T-learner, DR-learner and EP-learner can produce asymptotically unbiased estimates of the CATE by restricting their analyses to complete cases, and adjusting for $Z$ through their outcome models $\mu^a(Z)=\E[Y|A=a,C=1,Z]$, $A\in\{0,1\}$. We refer to this approach as an ``available case analysis'', and note that while the outcome regressions are limited to complete cases, the propensity score models should still be estimated using the full dataset. 

Available case analyses offer the simplest way of estimating the CATE in the presence of MAR outcome data, however, their asymptotic properties do not assure that they perform well when fit using finite samples. Instead, available case analyses can often be inefficient, as their outcome regressions restrict the population to those with complete cases, with information being ignored for individuals who have a missing outcome. Equally, when the outcome models are fit using ML, they will be prone to over-smoothing in the subsets of the population which experience high levels of outcome missingness. This can cause complex non-linear CATEs to be missed, with these approaches over-smoothing their outcome predictions due to the missing data, or can lead to estimators identifying heterogeneity where non-exists, for instance when the missingness only occurs within one treatment arm. Specific examples of data generating processes (DGPs) where this can occur are presented in Section \ref{Sims}. 

Because of these limitations, some authors choose to address missing outcome data by utilizing missing data techniques. A common missing data technique is to impute outcomes, replacing the missing outcomes with outcome predictions gained from an imputation model, $\hat{\E}[Y|C=1,A,Z]$ \citep{RN19,RN20}. These are easy to implement when using the existing non-parametric CATE estimators, as the estimators can be run using the imputed, complete dataset. However, as outcome imputations require the estimation of an additional nuisance model, plug-in bias can be introduced, with errors in the outcome imputations propagating through to the estimates of the CATE. Alternatively, other authors suggest addressing missing outcome data by re-weighting observed individuals based on their probability of having a non-missing outcome, $G(A,Z)=P[C=1|A,Z]$ \citep{RN1,RN26}. These weights can be used to estimate the CATE using a similar approach to the one \cite{RN10} uses to incorporate IPTWs, weighting observed individuals to create pseudo-outcomes, then regressing the pseudo-outcomes against $X$ to estimate the CATE. However, this estimator, which we will refer to as the IPTW-IPCW estimator, also suffers from plug-in bias, as errors in the IPTWs and IPCWs will propagate through to the CATE estimates. We therefore construct IF based estimators that incorporate these weights naturally, and which offer greater robustness to errors in the missingness predictions $G(A,Z)$.

\section{DR-learner/EP-learner extensions} \label{Method}
In this section, we extend the DR-learner \citep{RN10} and EP-learner \citep{RN50} to handle missing outcome data, leading to two new estimators, the mDR-learner (missing outcome DR learner) and mEP-learner (missing outcome EP learner) respectively.  

\subsection{mDR-learner} \label{DR_ext}
\begin{figure}[H]
\begin{center}
\includegraphics[scale=0.9]{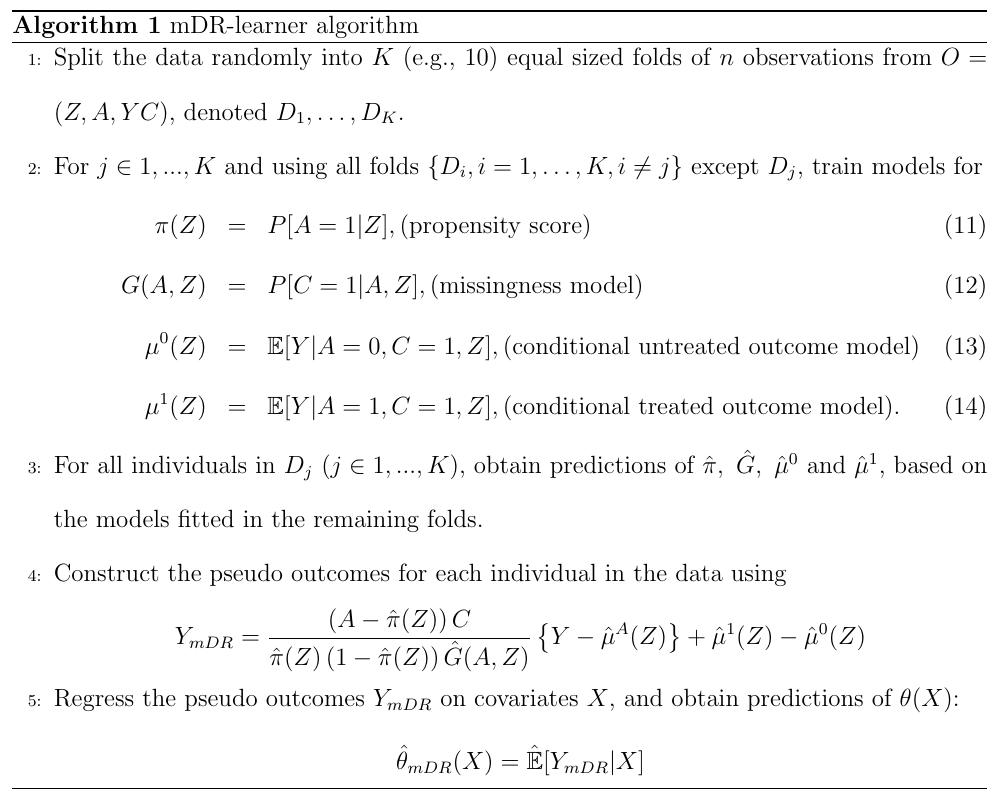} 
\end{center}
\caption{mDR-learner algorithm}
\label{alg:mDR}
\end{figure}

We begin by considering an extension of the DR-learner to the MAR outcome data setting. Recall that the DR-learner does not account for the under-representation that occurs as a result of MAR outcomes. This is because the DR-learner's pseudo-outcomes are derived using a risk function which assumes complete data. However, when outcomes are MAR, the risk function takes a new form, $\psi_{CATE}=\E[\theta(X)^2 -2\theta(X)(\E[Y|C=1,A=1,Z]-\E[Y|C=1,A=0,Z])]$, which now involves the indicator for outcomes being non-missing. It can be shown (Appendix \ref{EIF}) that the EIF is
\begin{equation}\label{MSE_IF_miss}
    \phi = \left(\theta^2(X) - 2\theta(X)\left(\mu^1(Z)-\mu^0(Z)\right)\right) - \frac{2\theta(X)  (A-\pi(Z))C}{\pi(Z)(1-\pi(Z))G(A,Z)}\{Y-\mu^A(Z)\} - \psi_{CATE}.    
\end{equation}
Using this EIF, we can construct pseudo-outcomes for the mDR-learner as:
\begin{equation} \label{ATE_EIF_MISS}
    Y_{mDR} = \frac{\left(A-\pi(Z)\right)C}{\pi(Z)\left(1-\pi(Z)\right)G(A,Z)}\left\{Y - \mu^{A}(Z)\right\} + \mu^1(Z) - \mu^0(Z).
\end{equation}
We note that IPCWs now appear in alongside IPTWs, allowing these pseudo-outcomes to account for the shift in covariate distribution caused by both missing outcome data and confounding. The mDR-learner then proceeds by regressing the estimates of these pseudo-outcomes $Y_{mDR}$ against covariates $X$ to obtain estimates of the CATE (See Algorithm \ref {alg:mDR}).

As the mDR-learner defines its pseudo-outcomes using the EIF of the MSE for the CATE (under outcome missingness), it not only minimizes the MSE risk function, but also experiences less sensitivity to errors in its nuisance functions (including the IPCWs). We demonstrate this by exploring the excess risk of the mDR-learner, defined as the difference in MSE risk function, $L(.)$, when evaluated at $\hat{\theta}(X)$ and at $\theta^*(X)$, where $L(\theta^*) = \inf_{\theta}L(\theta)$. Using the approach laid out by \cite{RN22}, we provide an upper bound for the excess risk (See Appendix \ref{App_ex_risk}). This bound demonstrates how the gradient of the MSE risk function for the mDR-learner has reduced sensitivity to errors in the nuisance functions, which now include the missingness model, $G(A,Z)$. It also shows how the mDR-learner can obtain oracle efficiency under similar assumptions to those of the DR-learner. 

For the mDR-learner to obtain oracle efficiency, it requires that two conditions hold; firstly, that the product of the convergence rates for the outcome predictions and propensity score estimates is faster than the rate for the oracle estimator, and secondly, that the product of the convergence rates for the outcome predictions and missingness estimates is also faster than the rate for the oracle estimator. For instance, consider an example where the oracle CATE estimator converges at a $\sqrt{n}$ rate. In this example, if we wish to estimate the nuisance functions within the mDR-learner without impacting the overall convergence rate of our CATE estimator, we require that the estimates of each of the nuisance functions converge at rates faster than $\sqrt[4]{n}$, hence allowing the two products of these rates to converge faster than $\sqrt{n}$. Equally, when the outcome predictions converge at rates slower than this, oracle efficiency can still be obtained as long as the propensity score estimates and missingness function estimates converge at rates which are sufficiently fast for the above conditions to hold. Under these weakened convergence requirements, ML can then be used when estimating nuisance functions within the mDR-learner without errors propagating through to the CATE.

\subsection{mEP-learner} \label{EP_ext}
We now demonstrate how the EP-learner, a targeted-learning based framework, can be extended for the setting with MAR outcome data. When outcomes are MAR, the EP-learner fails to account for under-representation introduced by missing outcomes. If we wish to account for this, we must recalculate the EIF which it uses to derive its pseudo-outcomes, with the new EIF taking a new form, eq. (\ref{MSE_IF_miss}). This EIF contains IPCWs within its drift term (second term), and hence, if we wish to set the sample average of the drift term to 0, removing plug-in bias, we must update the iTMLE process used within the EP-learner. 

To do so, we re-define the weight used in the iTMLE algorithm by considering eq. (\ref{MSE_IF_miss}), defining the weight to be $\hat{H}(A,C,Z)=\frac{CA}{\hat{G}(A,Z)\hat{\pi}(Z)} + \frac{C(1-A)}{\hat{G}(A,Z)\left(1-\hat{\pi}(Z)\right)}$. This weight ensures that the sample average of the second term in eq. (\ref{MSE_IF_miss}) converges to 0, and hence the mEP-learner will also be oracle efficient when its nuisance function estimates converge sufficiently fast. 

\begin{figure}[ht]
\begin{center}
\includegraphics[scale=0.9]{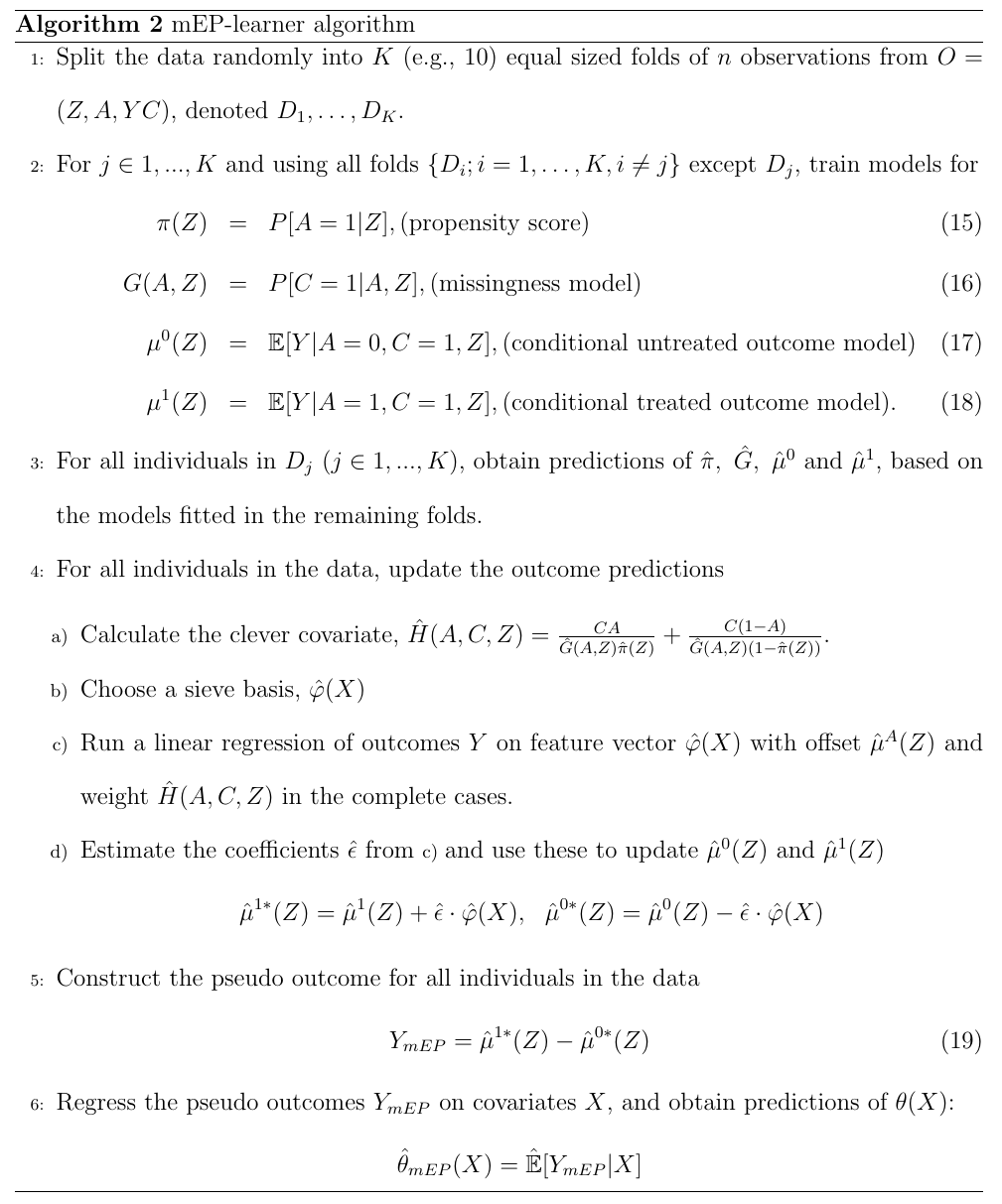} 
\end{center}
\caption{mEP-learner algorithm}
\label{alg:mEP}
\end{figure}

\subsection{Implementation} \label{implementation}

As both the mDR-learner and mEP-learner are general frameworks for estimating the CATE with MAR outcome data, their implementation requires the user to make several key decisions. In this section, we break these decisions down into two groups; (a) decisions required for obtaining CATE estimates, and (b) decisions required for assessing CATE performance. 

\subsubsection{Decisions required for obtaining CATE estimates} 
As both learners are model-agnostic the user must first decide how to estimate the nuisance functions/pseudo-outcome model. Data adaptive techniques can be chosen, however, if oracle performance is to be achieved, the estimates from the nuisance models must converge sufficiently fast to the truth. For this reason, we illustrate their implementation using the Super Learner \citep{RN49}, an ensemble learner that allows a range of data-adaptive algorithms to be implemented and which performs asymptotically as well as its best candidate learner.

After algorithm choice, users are required to decide which type of sample splitting they will implement within the mDR or mEP learners, ensuring the data used in the pseudo-outcome regression is i.i.d. Various options exist for achieving this, including K-fold cross-fitting (see Section \ref{Background_nomiss}) or independent sample splitting \citep{RN10}, however, estimators that allocate fully independent data for each nuisance/target model are typically less efficient, introducing finite sample bias by reducing samples sizes. For this reason, we demonstrate the mDR-learner and mEP-learner using a 10-fold cross-fitting process (see Algorithm \ref{alg:mDR}), but note that the appropriate number of folds will depend on the complexity and smoothness of the underlying nuisance/target functions. If users wish to explore variations of cross-fitting approaches, sensitivity analyses could be run using an alternative number of splits.

We also highlight that when cross-fitting is used, reduced training sample sizes for each model can introduce positivity violations, leaving certain subgroups poorly represented within some folds. To overcome this, the mDR-learner and mEP-learner can be run multiple times ($J$ times), using a different seed for the cross-fitting split. This results in a vector of CATE estimates for each individual, $\hat{\theta}_{s}(X)$, $s=1,..,J$. Final CATE estimates are then obtained by taking the median across these estimates, $\widehat{\theta}(X) = median\{\widehat{\theta}_{s}(X)\}$ \citep{RN21}. 

Finally, when implementing the mEP-learner, the sieve basis used within the iTMLE process must be specified. The existing iTMLE implementations use a univariate trigonometric cosine polynomial basis, as it offers strong approximation guarantees under smoothness assumptions \citep{RN54}. We also implement this sieve basis, following the guidance of \cite{RN54} to define the dimension of the sieve and its interaction order. However, alternative options exist, such as a cross-validation option for choosing the sieves tuning parameters \citep{RN50} and a penalized iTMLE implementation, which can improve performance for large sieves \citep{RN55}. 

\subsubsection{Decisions required for assessing CATE performance}
In addition to obtaining CATE estimates, users may also want to assess the accuracy of these estimates by obtaining measures of uncertainty or calculating evaluation metrics. Calculating measurements of uncertainty for CATE estimates generated using non-parametric estimators is challenging, as the theoretical convergence guarantees required for confidence intervals (CIs) to be derived are often not met. However, recent work by \cite{RN44,RN42} and \cite{RN62} offer potential solutions. In this work we explore how one of these techniques can be used with our estimators, focusing on the half-sample bootstrap approach \citep{RN62}, which can provide CIs when kernel based approaches are used to estimate the pseudo-outcome regression. Further details on this technique are outlined in Appendix \ref{APP_bootstrap}, and we evaluate the performance of this approach within our simulation study (See Section \ref{Sims}) in Appendix \ref{Bootstrap_results}.


Additionally, we note that evaluating the performance of CATE estimators when using real world data is challenging, with only one of the two potential outcomes, $Y(0)$ and $Y(1)$, ever observed for each individual. This means standard metrics such as the MSE cannot be calculated and used for estimator comparison, and equally makes deriving EIF based estimators for these metrics tricky, as by definition, the MSE metric and its EIF will be 0 at the true CATE. Alternatively, model performance can be assessed by reviewing the stability of model estimates across different sample splitting seeds, enabling users to understand the variability in individual CATE estimates. Additionally, if choosing between the mDR/mEP learners, users should focus on the strengths/weaknesses of each learner, with the mDR-learner performing best when the mEP-learner's sieve poorly approximates the CATE, i.e, when the CATE is non-smooth or sparse, while the mEP-learner may perform best when the mDR-learner's weights and pseudo-outcomes are highly variable/unstable.

\section{Simulation study}  \label{Sims}

\subsection{Set up}   \label{Sim_outline}
We study the finite sample performance of the mDR-learner and mEP-learner across three data generating processes (DGPs), where each DGP corresponds to a setting in which missing outcome data can complicate the estimation of the CATE. In each setting, we generate 6 uniformly distributed covariates $Z$, a binary treatment $A$ and a continuous outcome $Y$. In the first two DGPs, we define a simple unexposed outcome function $\mu^0(Z)$, a complex CATE $\theta(X)$ and define outcome missingness such that it occurs with high probability in only the treatment arm (DGP 1), or in both arms (DGP 2). This makes the complex CATE challenging to learn. In the third DGP, we define a complex unexposed outcome function, a simple CATE and define outcome missingness to occur with high probability in only the treatment arm, making the simple CATE difficult to learn (Appendix \ref{App_sim_specs}).  

We vary the training data sample size from 400 to 3200, and use 500 replicates for each scenario. We compare the mDR-learner and mEP-learner to four alternative CATE estimators; the IPTW-IPCW learner, the DR-learner, EP-learner and the T-learner, with the later three implemented using (1) available cases (Section \ref{Background_miss}) and (2) imputed outcomes. All estimators were implemented using 10 fold cross-fitting, with the nuisance models fit using the \textit{Super Learner} and the pseudo-outcome models fit using random forests (with 500 bootstrap half samples), enabling the generation of half-sample bootstrap CIs. To assess the performance of each estimator, we generated one test dataset with sample size $n=10,000$ per DGP and obtained the CATE estimates for each individual using each estimator. Performance was measured by calculating the root mean square \emph{median} error (RMSME) of each learner (Appendix \ref{App_sim_RMSME}), as mean root mean square error (RMSE) estimates were found to be skewed when using 500 replications. Conditional CI coverage was calculated/reported in Appendix \ref{Bootstrap_results}. For comparisons made using the mean RMSE, see Appendix \ref{App_sim}.  

\subsection{Findings}  \label{Sim_findings}

\begin{figure}[H]
\begin{center}
\includegraphics[scale=0.45]{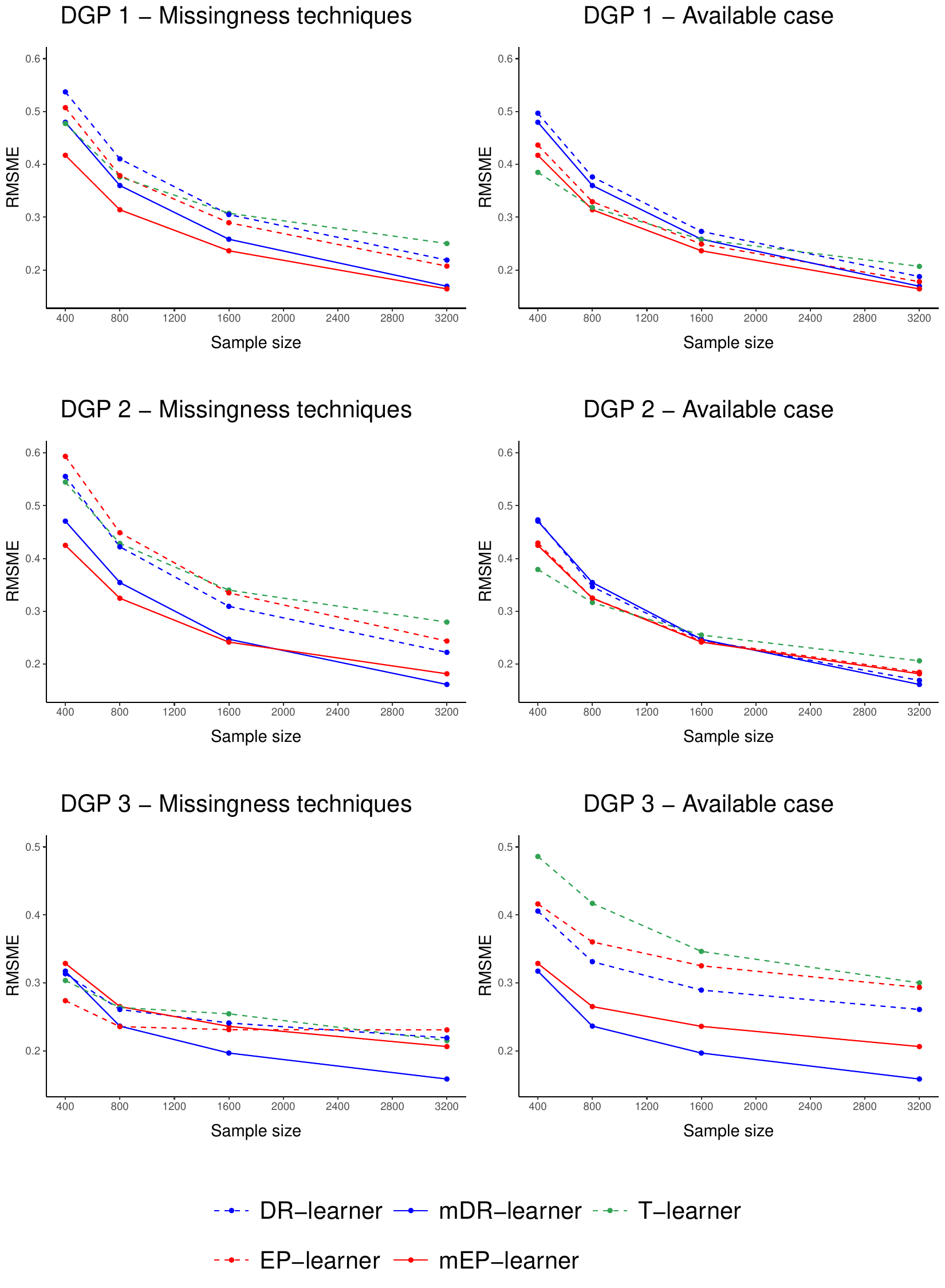} 
\end{center}
\caption{Root mean square median error (RMSME) for mDR-learner, mEP-learner, DR-learner, EP-learner and T-learner in three DGPs plotted by training sample size. Plots in the left column compare the mDR-learner and mEP-learner to the DR-learner, EP-learner and T-learner when used in combination with an outcome imputation model in DGP 1, 2 and 3 respectively. Plots in the right column compare the mDR-learner and mEP-learner to the available case versions of the DR-learner, EP-learner and T-learner in DGP 1, 2 and 3 respectively. }
\label{Sims_example_1_new}
\end{figure}

When comparing the mDR-learner and mEP-learner with the DR-learner and EP-learner using available cases (Figure \ref{Sims_example_1_new} - Right column), we see the mDR-learner and mEP-learner outperform the DR/EP learners respectively across all three DGPs. Equally, when comparing the mDR-learner and mEP-learner to the DR-learner and EP-learner fit using imputed outcomes (Figure \ref{Sims_example_1_new} - Left column) both learners outperformed their corresponding imputed outcome version when the CATE was complex (DGP 1 and 2), while when the CATE was simple, the imputed outcome DR-learner and EP-learner performed well. We also note how the IPTW-IPCW learner and available case/imputed outcome T-learner were sensitive to nuisance function complexity, with their performance depending heavily on the complexity of the outcome functions and propensity score/censoring functions respectively. IPTW-IPCW learner results are excluded from Figure \ref{Sims_example_1_new} to aid interpretability (See Appendix \ref{App_sim}).

Finally, we note how the CATE estimates obtained across simulations were more stable for the mEP-learner, EP-learner and T-learner compared to those obtained by the mDR-learner, DR-learner or IPTW-IPCW learner. This demonstrates how these estimators are prone to producing extreme CATE estimates when their weights are unstable. This can be seen more clearly when performance is measured using mean RMSE (Appendix \ref{App_sim}).

\section{GBSG2 trial analysis} \label{GBSG2}
\subsection{Background and methods} \label{ACTG175_back}
We illustrate the use of the mDR-learner and mEP-learner by applying them to the German breast cancer study (GBSG2) randomized trial \citep{RN70}. This trial randomly assigned patients to a hormonal therapy (n=440) or no hormonal therapy (n=246) after surgery and recorded baseline covariates on demographics, medical history and disease progression. Treatment efficacy was explored by recording recurrence free survival times, and we define our outcome of interest as breast cancer recurrence or death within three years of surgery. As some patients leave the study before making it to three years, missing outcome data is present, with 158 (46.5\%) and 66 (26.8\%) of the randomized patients lost to follow-up in each treatment arm. We conduct an intention-to-treat analysis and estimate two CATEs: one conditioned on all baseline covariates and one conditioned solely on progesterone receptor levels (fmol/l), where higher levels are associated with greater benefits from hormonal therapies. 

In this trial, patients with non-missing outcomes had higher average progesterone receptor levels at baseline than the full randomized population, with a greater increase seen in amongst patients in the hormonal therapy arm. If left unaccounted for, this may result in CATE estimates which suggest hormonal therapies have a greater benefit than is true. Instead, we estimate CATEs using the mDR-learner and mEP-learner, and compare these to estimates from the the DR-learner, EP-learner, T-learner and IPTW-IPCW learner, with the first three fit using available case analyses as well as in combination with imputed outcomes. All nuisance models were fit using all baseline covariates and all models, including the pseudo-outcome models were estimated using a \textit{Super Learner}, with the focus on obtaining accurate point estimates rather than CIs. The DR, EP and IPTW-IPCW learners are implemented using 10 fold cross-fitting and we report the median CATE estimate over 10 different sample splitting random seeds. Estimates with CIs that were obtained using untuned random forests (and 500 bootstrap samples) are reported in Appendix \ref{APP_GB_RF}.

\subsection{Findings} \label{ACTG_findings}

\begin{figure}[H]
\begin{center}
\includegraphics[scale=0.45]{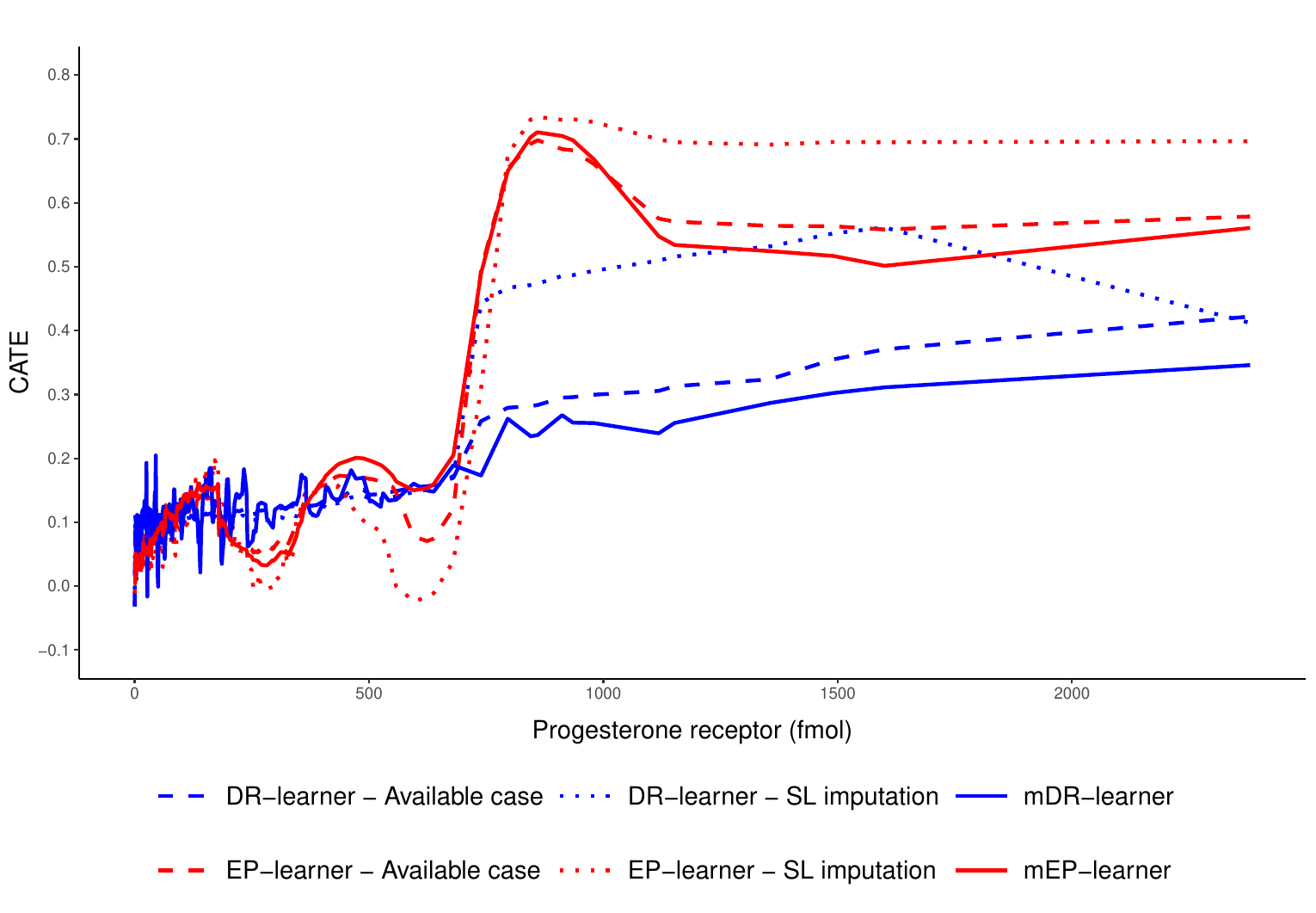} 
\end{center}
\caption{Median CATE estimates plotted by progesterone receptor (fmol/l).}
\label{Data_ex_1}
\end{figure}

\begin{figure}[H]
\begin{center}
\includegraphics[scale=0.45]{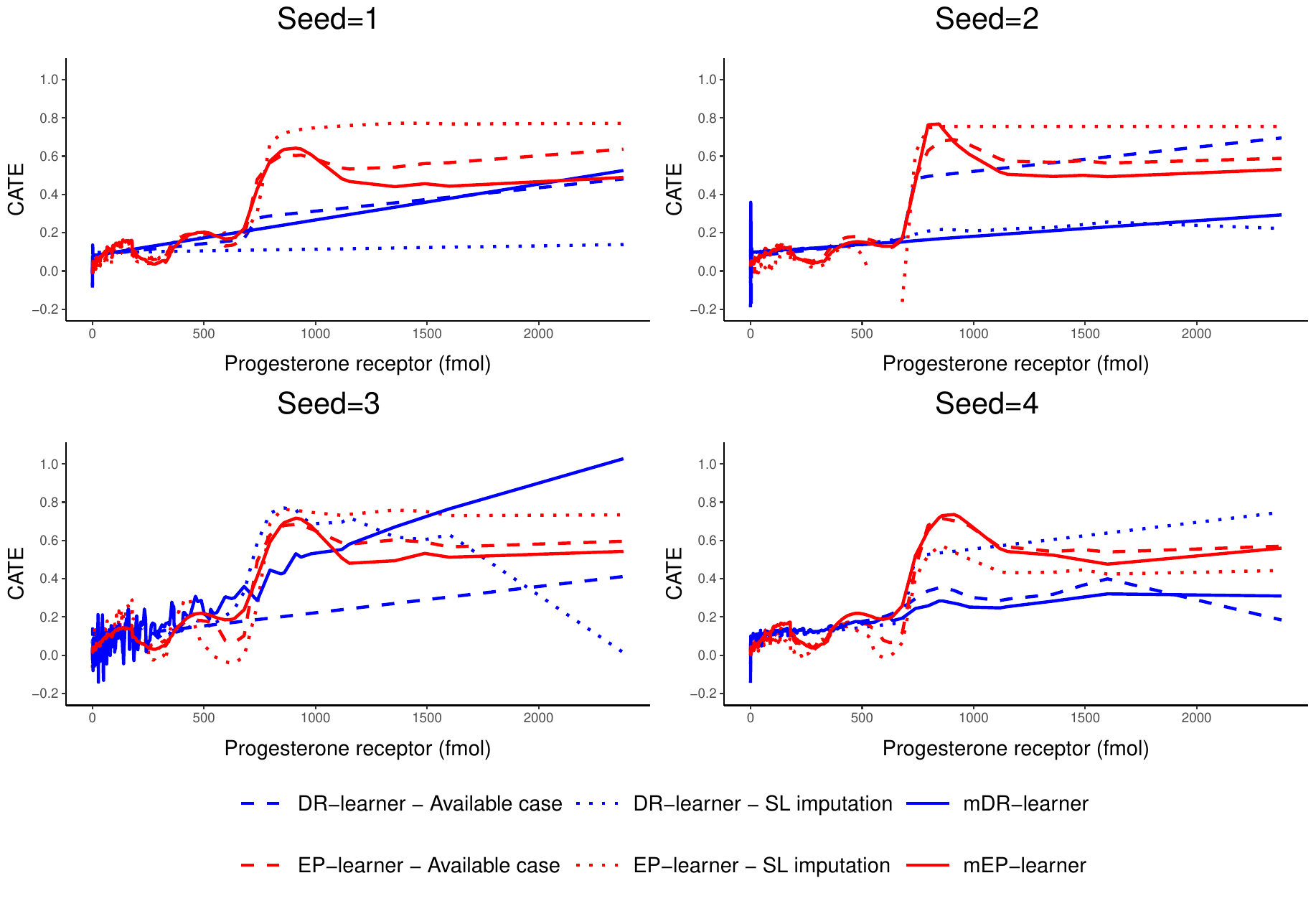} 
\end{center}
\caption{CATE estimates from single cross-fitting seeds plotted by progesterone receptor (fmol/l).}
\label{Data_ex_2}
\end{figure}

Figure \ref{Data_ex_1} shows the median CATE estimates conditional on progesterone receptor levels only. The available case DR and EP learners produce higher CATE estimates than the mDR and mEP learners respectively. This is expected as the observed hormonal therapy arm has higher progesterone receptor levels, and once adjusted for, we would expect to see smaller CATE estimates. Additionally, we note that the estimates from the DR and EP learners fit using imputed outcomes increase rather than decrease. We also review the CATE estimates which are conditional on all baseline covariates, with Table \ref{Data_ex_tab_1} reporting the mean CATE estimates for individuals in five progesterone receptor groups. Similar trends are seen in ``1500-1999" receptor level category, however, trends are less obvious in areas of the population with good representation. Table \ref{Data_ex_tab_1} also reports estimates from the IPTW-IPCW learner, which are highly unstable, and estimates from the T-learner, which suggest smaller treatment effects. Finally, we note that greater stability is observed for estimates obtained from the EP-learner variations than the DR-learner variations (Figure \ref{Data_ex_2}), highlighting the DR/mDR-learner's instability.

\section{Discussion}  \label{Discussion}

In this paper, we discussed the commonly overlooked problem of estimating the CATE when outcome data is MAR. Our work proposes two robust oracle efficient estimators, the mDR-learner and mEP-learner, which address the population imbalances introduced by missing outcome data by robustly incorporating IPCWs into the DR-learner and EP-learner respectively. Our proposed approaches and implementation guidance have the potential to help improve CATE estimation in real-world data settings where outcome data is MAR. However, there remains considerable scope for further developments in this area.

Firstly, the existing tools for obtaining CIs for non-parametric CATEs either restrict the form of the CATE or limit the estimation tools which can be used. We think further development of these techniques to allow for a wider variation of estimation techniques would greatly improve the utility of these approaches. Additionally, when generating half sample bootstrap CIs, we observed very poor coverage for certain individuals, along with very wide CIs intervals for others. For these CIs to have utility in practical examples, improved conditional coverage will be required. 

Finally, we think there is great scope for further extensions of these techniques to handle more complex data, e.g., post-baseline covariate information or missing covariate data. We outline an example of one of these extensions in Appendix \ref{Panel_ext}, where we discuss how the mDR-learner could be extended to handle post-baseline covariates. 

\newpage

\newpage
\begin{center}
\captionsetup{type=table}  
\caption{Mean CATE estimates by progesterone receptor groups when allowing the CATE to be conditional on all baseline covariates. } 
\begin{sideways} \label{Data_ex_tab_1}
\begin{tabular}{cccccccccccccccc} 
\hline
\multicolumn{1}{c}{Progesterone receptor} &  & \multicolumn{1}{c}{mDR}  &  & \multicolumn{1}{c}{mEP}  &  & \multicolumn{2}{c}{DR}   &  & \multicolumn{2}{c}{EP}   &  & \multicolumn{2}{c}{T} &  & \multicolumn{1}{c}{IPTW-IPCW} \\
 (Categorical) &  &  &  &  &  & \multicolumn{1}{c}{AC} & \multicolumn{1}{c}{Imputation} &  & \multicolumn{1}{c}{AC} & \multicolumn{1}{c}{Imputation} &  & \multicolumn{1}{c}{AC} & \multicolumn{1}{c}{Imputation} &  &    \\\hline
$<500$       &  & 0.095   &  & 0.080    &  & 0.114  & 0.109  & & 0.081 & 0.067 & & 0.075 & 0.061 & & 0.069\\
$500-999$    &  & 0.051   &  & 0.397    &  & -0.074 & -0.084 & & 0.402 & 0.157 & & 0.102 & 0.149 & & -0.785\\
$1000-1499$  &  & 0.388   &  & 0.405    &  & 0.401  & 0.393  & & 0.408 & 0.345 & & 0.292 & 0.282 & & 0.144\\
$1500-1999$  &  & 0.452   &  & 0.336    &  & 0.502  & 0.517  & & 0.358 & 0.285 & & 0.333 & 0.304 & & 0.857\\
$\geq 2000$  &  & 0.290   &  & 0.239    &  & 0.283  & 0.282  & & 0.244 & 0.195 & & 0.179 & 0.166 & & 0.299\\
 \hline
 \multicolumn{15}{l}{\small\textit{AC = Available case}}
\end{tabular}
\end{sideways}
\end{center}

\section*{Acknowledgements}
This work was supported by the Medical Research Council [grant number MR/N013638/1]. KDO was funded by a Royal Society-Welcome Trust Sir Henry Dale fellowship, grant number 218554/Z/19/Z. RHK was funded by UK Research and Innovation (Future Leaders Fellowship MR/X015017/1). SV was supported by Advanced ERC grant ACME (101141305).

\section*{Supplementary Materials}

Web Appendices referenced throughout are available with this paper at the Biometrics website on Oxford Academic, and the code used to implement the learners described in this paper can be found at \url{https://github.com/Matt-Pryce/mDR-learner_mEP-learner}. 

\section*{Data Availability}
The GBSG2 dataset used in this paper can be accessed through the TH.data package in R.

\bibliographystyle{unsrtnat}

\bibliography{Paper_endnote_lib.bib}

\newpage
\appendix

\section{Illustration of T-learner bias} \label{ill_ex_CC}

We illustrate, through simulated data, how the T-learner, which is constructed by optimizing predictions for conditional mean outcome functions in the treated and untreated arms, does not deliver the correct bias-variance trade-off. 

Consider three covariates $\{Z_1,Z_2,Z_3\}$, generated uniformly between -1 and 1. Then let $A$ be a binary treatment from a Bernoulli distribution with probability of $\frac{1+sin(Z_1)}{2}$, and $Y$ be a binary outcome $Y$ with probability:
$$\frac{1}{1+e^{-\left(\frac{4Z_1^2}{2}\cdot\mathbbm{1}(Z_1)-Z_1\cdot(1-\mathbbm{1}(Z_1))+Z_1\cdot(1-\mathbbm{1}(Z_1))+\frac{Z_2}{2}+\frac{Z_3}{2}\right)}}$$
\noindent where
\begin{align*}
 \mathbbm{1}(Z_1)=\begin{cases} 
      0 & x < -0.5 \\
      1 & -0.5 \leq x \leq 0.5 \\
      0 & 0.5 < x .
   \end{cases}
\end{align*}
Note that in this example no treatment effect exists, i.e. the difference between the two functions $\mu^1$ and $\mu^0$  is a simple function (a constant, equal to zero), but the individual functions are complex and therefore difficult to estimate. 

\begin{figure}[ht]
\begin{center}
\includegraphics[scale=0.55]{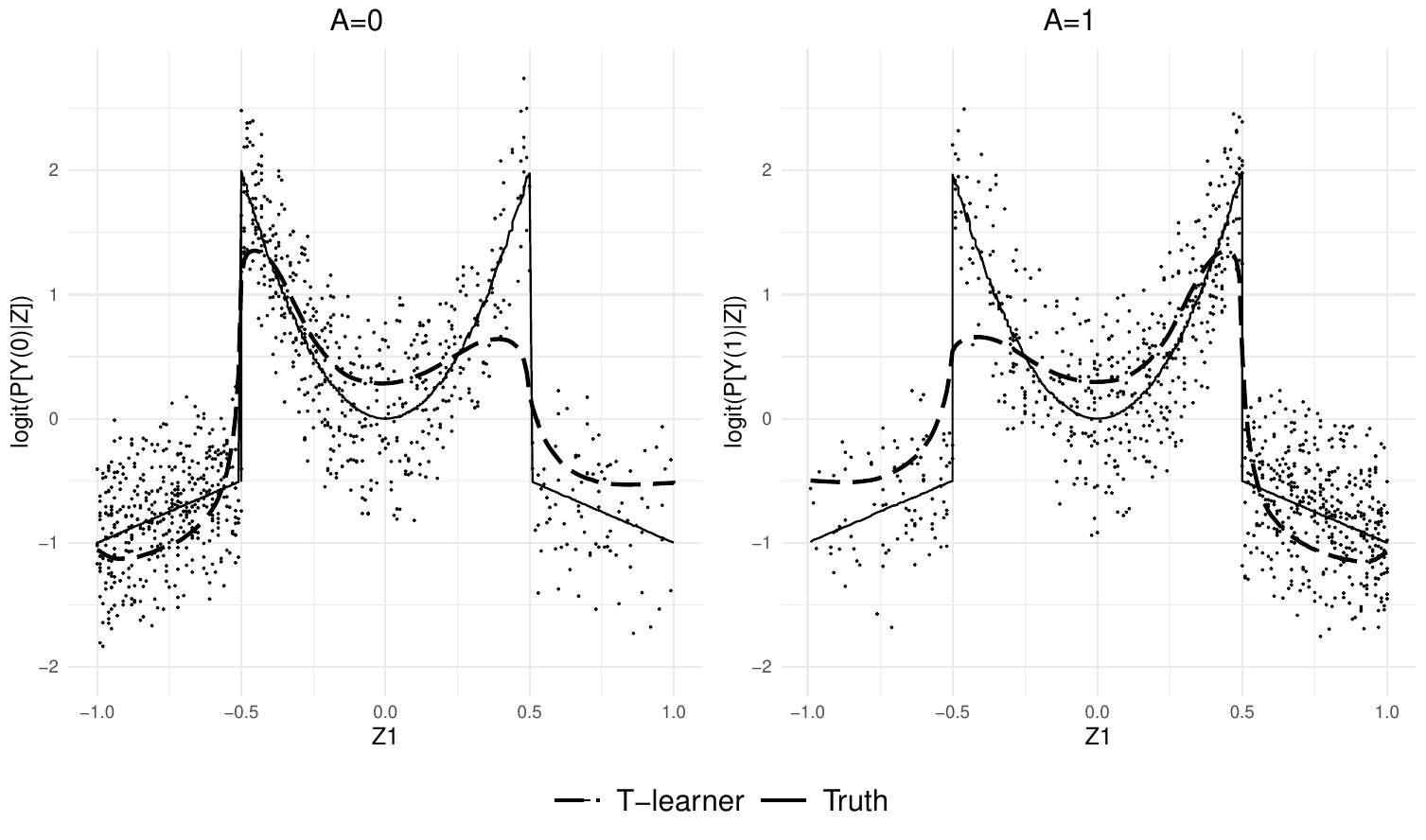} 
\end{center}
\caption{Simulated data example plotting the logit of the outcome probabilities in each treatment arm across $Z_1$ values for 2000 data points. Dashed line presents the estimates made by the T-learner.}
\label{No_miss_plot_1}
\end{figure}

Figure \ref{No_miss_plot_1} presents 2000 data points from this example, with the true $\mu^1$ and $\mu^0$, i.e. the logit of the potential outcome probabilities plotted against their $Z_1$ values, for untreated and treated individuals. 

In this setting, the untreated individuals with $Z_1$ values above 0.5 are under-represented, while treated individuals with $Z_1$ values less than -0.5 are also under-represented. As a result, the outcome models for the treated/untreated are more difficult to fit in those regions, and the machine learning models needed to fit a T-learner will over-smooth in the respective areas of $Z_1$ which are under-represented. Figure \ref{No_miss_plot_1} demonstrates this by plotting the estimated $\hat{\mu^a}(Z)$ obtained using tuned random forests, across 500 simulations, which are later used to construct a T-learner. 

\begin{figure}[ht]
\begin{center}
\includegraphics[scale=0.5]{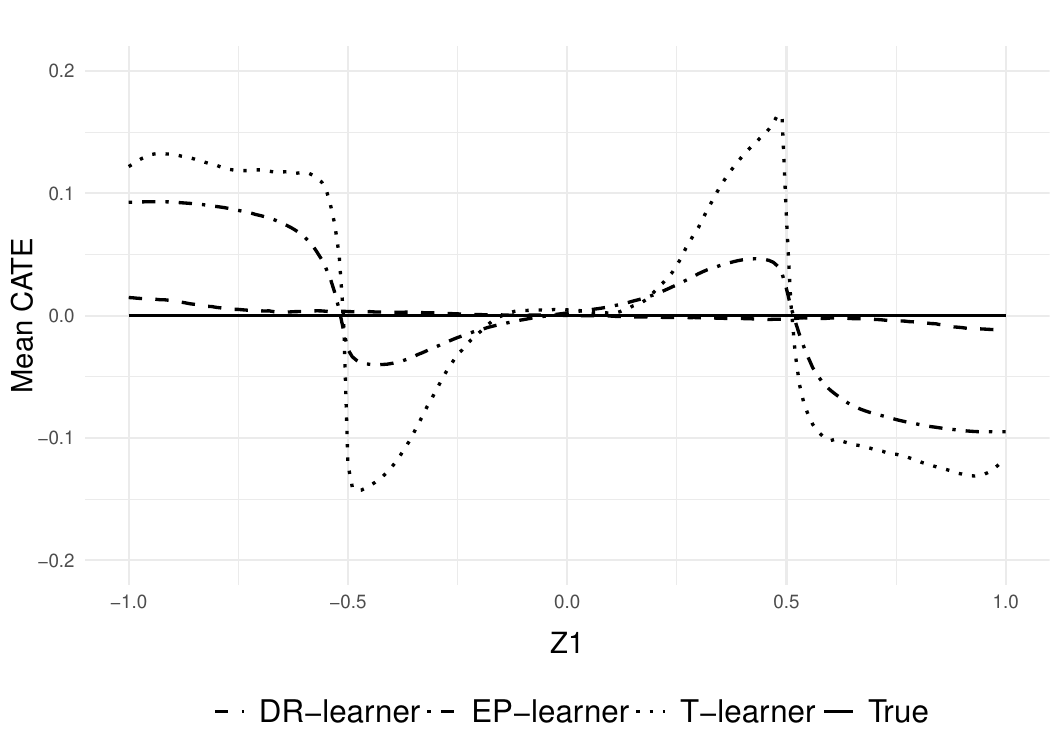} 
\end{center}
\caption{Mean of CATE estimates across 500 simulations, plotted by $Z_1$ values for the T-learner, DR-learner and EP-learner.}
\label{No_miss_plot_2}
\end{figure}

As a consequence of the uneven smoothing across the treated/untreated groups, the resulting plug-in CATE estimates are volatile and heavily prone to bias (Figure \ref{No_miss_plot_2}). In contrast, as the DR-learner and EP-learner target the CATE directly, they express less volatility.

\newpage
\section{DR-learner algorithm} \label{App_DR}
\begin{algorithm}
\caption{DR-learner algorithm}\label{alg:DR}
\begin{algorithmic}[1] 
\State Split the data randomly into $K$ (e.g., 10) equal sized folds of $n$ observations from $O = (Z, A, Y)$. Denote  $D_{1},\ldots,D_{K}$.
\State For $j \in 1,...,K$ and using all folds $\{D_{i},i=1,\dots,K, i\neq j\}$ except $D_{j}$, train models for  
\begin{eqnarray}
\pi(Z)&=&P[A=1|Z], \text{(propensity score)}\\
\mu^0(Z)&=&\E[Y|A=0,Z], \text{(conditional untreated outcome model)} \\
\mu^1(Z)&=&\E[Y|A=1,Z],  \text{(conditional treated outcome model)}.
\end{eqnarray}
\State For all individuals in $D_{j}$ ($j \in 1,...,K$), obtain predictions of $\hat{\pi}, ~\hat{\mu}^0$ and $\hat{\mu}^1$, based on the models fitted in the remaining folds.
\State Construct the pseudo outcomes for each individual in the training data using $$Y_{DR} = \frac{A-\hat{\pi}(Z)}{\hat{\pi}(Z)\left(1-\hat{\pi}(Z)\right)}\left\{Y - \hat{\mu}^{A}(Z)\right\} + \hat{\mu}^1(Z) - \hat{\mu}^0(Z)$$
\State Regress the pseudo outcomes $Y_{DR}$ on covariates $X$, and obtain predictions of $\theta(X)$:  
$$ \hat{\theta}_{DR}(X) = \hat{\E}[Y_{DR}|X]$$
\end{algorithmic}
\end{algorithm}

\newpage
\section{EP-learner algorithm} \label{App_EP}
\begin{algorithm}
\caption{EP-learner algorithm}\label{alg:EP}
\begin{algorithmic}[1] 
\State Split the data randomly into $K$ (e.g., 10) equal sized folds of $n$ observations from $O = (Z, A, Y)$. Denote  $D_{1},\ldots,D_{K}$.
\State For $j \in 1,...,K$ and using all folds $\{D_{i},i=1,\dots,K, i\neq j\}$ except $D_{j}$, train models for  
\begin{eqnarray}
\pi(Z)&=&P[A=1|Z], \text{(propensity score)}\\
\mu^0(Z)&=&\E[Y|A=0,C=1,Z], \text{(conditional untreated outcome model)} \\
\mu^1(Z)&=&\E[Y|A=1,C=1,Z],  \text{(conditional treated outcome model)}.
\end{eqnarray}
\State For all individuals in $D_{j}$ ($j \in 1,...,K$), obtain predictions of $\hat{\pi},~\hat{G}, ~\hat{\mu}^0$ and $\hat{\mu}^1$, based on the models fitted in the remaining folds.
\State  Using all individuals in the data, update outcome predictions
\begin{enumerate}
  \item[{\small a)}] Construct the clever covariate, $\hat{H}(A,Z)=\frac{A}{\hat{\pi}(Z)} + \frac{1-A}{\left(1-\hat{\pi}(Z)\right)}$. 
  \item[{\small b)}] Choose a sieve basis, $\hat{\varphi}(X)$
  \item[{\small c)}] Run a linear regression of outcomes $Y$ on feature vector $\hat{\varphi}(X)$ with offset $\hat{\mu}^A(Z)$ and weight $\hat{H}(A,Z)$ in the complete cases. 
  \item[{\small d)}] Estimate the coefficients $\hat{\epsilon}$ from \small{c)} \normalsize and use these to update $\hat{\mu}^0(Z)$ and $\hat{\mu}^1(Z)$
$$\hat{\mu}^{1*}(Z)=\hat{\mu}^{1}(Z)+\hat{\epsilon}\cdot\hat{\varphi}(X),~~\hat{\mu}^{0*}(Z)=\hat{\mu}^{0}(Z)-\hat{\epsilon}\cdot\hat{\varphi}(X)$$
\end{enumerate}
\State  Construct the pseudo outcome for all individuals in the training data
\begin{equation} \label{pse_eq}
    Y_{EP} = \hat{\mu}^{1*}(Z) - \hat{\mu}^{0*}(Z)
\end{equation}
\State Regress the pseudo outcomes $Y_{EP}$ on covariates $X$, and obtain predictions of $\theta(X)$:  
$$ \hat{\theta}_{EP}(X) = \hat{\E}[Y_{EP}|X]$$
\end{algorithmic}
\end{algorithm}


\section{Derivation of the EIF used within the mDR-learner and mEP-learner} \label{EIF}
Let us define $\mathcal{P}$ as the true observed data distribution and let our estimand of interest be
$$\psi(\mathcal{P}) = \E_{\mathcal{P}}[\theta(X)^2 - 2\theta(X)(\E_{\mathcal{P}}(Y(1)|X) - \E_{\mathcal{P}}(Y(0)|X))],$$ 
for a given $\theta(x)$. Assuming A1-A5 hold, we can write this risk function as:
\begin{align*}
\psi(\mathcal{P}) = \E_{\mathcal{P}}\left[\theta^2(X) - 2\theta(X)(\E_{\mathcal{P}}[Y|A=1,C=1,Z] - \E_{\mathcal{P}}[Y|A=0,C=1,Z])\right]
\end{align*}

\noindent Expanding this, we can write our estimand as
\begin{gather}\label{mEP_estimand}
\psi(\mathcal{P}) = \int f(z)\left(\theta^2(x) - 2\theta(x)(\E_{\mathcal{P}}[Y|A=1,C=1,Z=z]-\E_{\mathcal{P}}[Y|A=0,C=1,Z=z])\right) dz 
\end{gather}

\noindent
To derive the efficient influence function for (\ref{mEP_estimand}), we perturb $\mathcal{P}$ in the direction parameterized via the one-dimensional mixture model
$$\mathcal{P}_t = t\Tilde{\mathcal{P}} + (1-t)\mathcal{P},$$
where $\Tilde{\mathcal{P}}$ a fixed,
deterministic distribution with its support contained in the support of $\mathcal{P}$. By perturbing $\mathcal{P}$
in the direction of a point mass at $(\Tilde{z},\Tilde{x},\Tilde{a},\Tilde{c},\Tilde{y})$, we get:
\begin{gather*}
\psi(\mathcal{P}_t) = \int f_{\mathcal{P}_t}(z)\left(\theta^2(x) - 2\theta(x)\left(\E_{\mathcal{P}_t}[Y|A=1,C=1,Z=z]- \E_{\mathcal{P}_t}[Y|A=0,C=1,Z=z]\right)\right) dz
\end{gather*}
\noindent where, 
\begin{align*}
    f_{\mathcal{P}_t}(z) = t\mathbbm{1}_{\Tilde{z}}(z) + (1-t)f(z) 
\end{align*}
\noindent We can then derive the EIF by calculating the Gateaux derivative:
\begin{allowdisplaybreaks}
\begin{align*}
\frac{d\psi(\mathcal{P}_t)}{dt}\Bigr|_{t=0} &= \int \frac{d}{dt}\left\{f_{\mathcal{P}_t}(z)\left(\theta^2(x) - 2\theta(x)(\E_{\mathcal{P}_t}[Y|A=1,C=1,Z=z]- \right.\right.\\ 
&~~~~~~~~~~~~~~~~~~~~~~~~~~~~~~~~~~~~~~~~~\left.\left. \E_{\mathcal{P}_t}[Y|A=0,C=1,Z=z])\right)\right\}\Bigr|_{t=0}dz \\
&= \int \left\{(\mathbbm{1}_{\Tilde{z}}(z) - f(z))\left(\theta^2(x) - 2\theta(x)(\E_{\mathcal{P}}[Y|A=1,C=1,Z=z]- \right.\right.\\ 
&~~~~~~~~~~~~~~~~~~~~~~~~~~~~~~~~~~~~~~~~~~~~~~~~\left.\left. \E_{\mathcal{P}}[Y|A=0,C=1,Z=z])\right) + \right. \\
&~~~~~~~\left.f(z)\frac{d}{dt}\left\{\theta^2(x) - 2\theta(x)(\E_{\mathcal{P}_t}[Y|A=1,C=1,Z=z]- \right.\right.\\
&~~~~~~~~~~~~~~~~~~~~~~~~~~~~~~~~~~~~~\left.\left.\E_{\mathcal{P}_t}[Y|A=0,C=1,Z=z])\right\}\right\}|_{t=0} dz\\
&= \int \left\{(\mathbbm{1}_{\Tilde{z}}(z) - f(z))\left(\theta^2(x) - 2\theta(x)(\E_{\mathcal{P}}[Y|A=1,C=1,Z=z]- \right.\right.\\ 
&~~~~~~~~~~~~~~~~~~~~~~~~~~~~~~~~~~~~~~~~~~~~~~~~\left.\left. \E_{\mathcal{P}}[Y|A=0,C=1,Z=z])\right) + \right. \\
&f(z)\left(- 2\theta(x)\frac{d}{dt}(\E_{\mathcal{P}_t}[Y|A=1,C=1,Z=z]- \E_{\mathcal{P}_t}[Y|A=0,C=1,Z=z])|_{t=0}\right)\}dz\\
&=  \int \left\{(\mathbbm{1}_{\Tilde{z}}(z) - f(z))\left(\theta^2(x) - 2\theta(x)(\E_{\mathcal{P}}[Y|A=1,C=1,Z=z]- \right.\right.\\ 
&~~~~~~~~~~~~~~~~~~~~~~~~~~~~~~~~~~~~~~~~~~~~~~~~\left.\left. \E_{\mathcal{P}}[Y|A=0,C=1,Z=z])\right) + \right. \\
&~~~~~~~~~f(z)\left(- 2\theta(x)\frac{\mathbbm{1}_{\Tilde{z}}(z)(\mathbbm{1}_{\Tilde{a}}(a)-\pi(z))\mathbbm{1}_{\Tilde{c}}(c)}{f(z)(1-\pi(z))\pi(z)G(a,z)}\{\Tilde{y}-\mu_A(z)\}\right)\}dz \\
&= \theta(\Tilde{x})^2 - 2\theta(\Tilde{x})(\mu^1(\Tilde{z}) - \mu^0(\Tilde{z}))  - \\
&~~~~2\theta(\Tilde{x})\frac{(\mathbbm{1}_{\Tilde{a}}(a)-\pi(\Tilde{z}))\mathbbm{1}_{\Tilde{c}}(c)}{(1-\pi(\Tilde{z}))\pi(\Tilde{z})G(\Tilde{a},\Tilde{z})}\{\Tilde{y}-\mu_A(\Tilde{z})\} - \psi(\mathcal{P})
\end{align*}
\end{allowdisplaybreaks}

\noindent where $\pi(z) = f(A=1|Z=z) = P(A=1|Z=z)$, $G(a,z) = f(C=1|A=a,Z=z) = P(C=1|A=a,Z=z)$ and $\mu^{a}(z) = \E_{\mathcal{P}}(Y|A=a,C=1,Z=z)$

\section{mDR-learner excess risk bound - Overview} \label{App_ex_risk}

To demonstrate that the estimation error of the nuisance functions only has a second-order impact and therefore the mDR-learner is approximately insensitive to this, we provide an upper bound for the excess risk of the mDR-learner. We do so by following the structure and Theorems laid out by \cite{RN22}. 

We define the excess risk as the difference 
\begin{equation}\label{ex-risk}
    L(\hat{\theta})-L(\theta^*)
\end{equation}
in MSE risk $L(.)$ when evaluated at $\hat{\theta}$ and at $\theta^*$, where $\theta^*$ is defined as the minimiser of risk function when the infimum of the risk function is obtained, $L(\theta^*) = \inf_{\theta}L(\theta)$. 

Let us also define $\Theta$ be the class of target parameter values, and $\hat{\Theta}$ to be the class of target parameter estimates, containing all possible $\hat{\theta}$. As the population risk function depends not only on $\theta\in\Theta$, but on a set of nuisance parameters $g=(\pi,G,\mu^0,\mu^1)$, we introduce to notation $L(\theta,g)$. Additionally, as the objective of the mDR-learner is to minimize the MSE risk function under known nuisance functions $g_0$, we write the excess risk that we wish to bound as:
\begin{equation}\label{ex-risk2}
    L(\hat{\theta},g_0)-L(\theta^*,g_0), 
\end{equation}
where $L(\theta^*,g_0) = \inf_{\theta}L(\theta,g_0)$. When the nuisance functions are known, an upper bound for this excess risk can be written in terms of the convergence rate of the CATE estimates obtained from mDR-learner's second stage pseudo-outcome regression: 
$$L(\hat{\theta},g_0)-L(\theta^*,g_0) \leq \text{Rate}_{\theta}(g_0). $$
In practice, the mDR-learner does not know the true nuisance parameters $g_0$, but estimates them, $\hat{g}$, with this upper bound taking the form:
$$L(\hat{\theta},\hat{g})-L(\theta^*,\hat{g}) \leq \text{Rate}_{\theta}(\hat{g}). $$

Below, we use the criteria outlined by \cite{RN22} to write the right hand side of this bound in terms of the known nuisance functions, rather that the estimated nuisance functions. We begin by defining the nuisance/target algorithms and their respective rates. 

\vspace{0.3cm}
\noindent \textbf{Notation:} Norms written $\|.\|_p$ refer to $l_p$ norm. 

\vspace{0.5cm}
\noindent \textbf{Definition 1} (Algorithms \& Rates). 
\begin{itemize}
    \item[\textbf{a)}] Nuisance algorithms and rates. \textit{For the mDR-learner, as $g=(\pi,\mu^0,\mu^1,G)$, we define the first stage algorithms and their corresponding rate functions for each component of $g$. }
    \item[] \textit{The algorithm} $\text{Alg}_{\pi}$\textit{ outputs a predictor of $\pi$, $\hat{\pi}$, for which}
\end{itemize}
\begin{equation*} \label{Def1a1}
    \|\hat{\pi}-\pi_0\|_2 \leq \text{Rate}_{\pi}(\delta)~~~~~with~probability~almost~1
\end{equation*}
\begin{itemize}
    \item[] \textit{The algorithm }$\text{Alg}_{G}$\textit{ outputs a predictor of $G$, $\hat{G}$, for which}
\end{itemize}
\begin{equation*}\label{Def1a2}
    \|\hat{G}-G_0\|_2 \leq \text{Rate}_{G}(\delta)~~~~~with~probability~almost~1
\end{equation*}
\begin{itemize}
    \item[] $\textit{The algorithm }\text{Alg}_{\mu^0}$\textit{ outputs a predictor of $\mu^0$, $\hat{\mu}^0$ for which}
\end{itemize}
\begin{equation*} \label{Def1a3}
    \|\hat{\mu}^0-\mu^0_0\|_2 \leq \text{Rate}_{\mu^0}(\delta)~~~~~with~probability~almost~1
\end{equation*}
\begin{itemize}
    \item[] \textit{The algorithm }$\text{Alg}_{\mu^1}$\textit{ outputs a predictor of $\mu^1$, $\hat{\mu}^1$ for which}
\end{itemize}
\begin{equation*}\label{Def1a4}
    \|\hat{\mu}^1-\mu^1_0\|_2 \leq \text{Rate}_{\mu^1}(\delta)~~~~~with~probability~almost~1
\end{equation*}

\vspace{0.1cm}
\begin{itemize}
    \item [\textbf{b)}] Target algorithm and rate. \textit{Let $\theta^*\in\hat{\Theta}$ and define the second stage algorithm} $\text{Alg}(g)$\textit{, with $g=(\pi,\mu^0,\mu^1,G)$, to be a predictor of $\theta$, for which}
\end{itemize}
\begin{equation*} \label{def1b}
    L(\hat{\theta},g) - L(\theta^*,g) \leq \text{Rate}_{\theta}(\delta;g)~~~~~with~probability~almost~1
\end{equation*}

\vspace{0.3cm}
\noindent \textbf{Notation:} When writing these rates we drop $\delta$ from the notation, writing $Rate_{\pi}$, $Rate_{\mu^0}$, $Rate_{\mu^1}$, $Rate_{G}$, $Rate_{\theta}(g)$.  

\vspace{0.5cm}
\noindent \textbf{Definition 2} (Directional Derivative). \textit{Let $\mathcal{F}$ be a vector space of functions. For a functional $F:\mathcal{F}\rightarrow \mathbbm{R}$, we define the derivative operator $\left.D_{f}F(f)[h]=\frac{d}{dt}F(f+th)\right|_{t=0}$ for a pair of functions $f,h\in\mathcal{F}$. Likewise, we define $\left.D^k_f F(f)[h_1,...,h_k]=\frac{\partial^k}{\partial t_1...\partial t_k}F(f+t_1h_1+...+t_kh_k)\right|_{t_1=...t_k=0}$}. \textit{When considering a functional in two arguments, e.g. $L(\theta,g)$, we write $D_{\theta}L(\theta,g)$ and $D_gL(\theta,g)$ to highlight which component the argument is taken with respect to.} 

\vspace{0.5cm}
We then define $\hat{\mathcal{G}}$ to be the class of nuisance parameter estimates and define the four assumptions needed to identify an excess risk upper bound that is written in terms of known nuisance functions:\\

\noindent \textbf{Assumption 1} (Orthogonal Loss). The population risk $L(\theta^*,g_0)$ is Neyman orthogonal:
\begin{equation} \label{A1.1_miss}
    D_\pi D_{\theta}L(\theta^*,g_0)[\theta-\theta^*,\pi-\pi_0] = 0 ~~~\forall \theta \in \hat{\Theta},\pi\in\hat{\mathcal{G}},
\end{equation}
\begin{equation} \label{A1.2_miss}
    D_{\mu^1} D_{\theta}L(\theta^*,g_0)[\theta-\theta^*,\mu^1-\mu^1_0] = 0 ~~~\forall \theta \in \hat{\Theta},\mu^1\in\hat{\mathcal{G}}, 
\end{equation}
\begin{equation} \label{A1.3_miss}
    D_{\mu^0} D_{\theta}L(\theta^*,g_0)[\theta-\theta^*,\mu^0-\mu^0_0] = 0 ~~~\forall \theta \in \hat{\Theta},\mu^0\in\hat{\mathcal{G}}, 
\end{equation}
\begin{equation} \label{A1.4_miss}
    D_G D_{\theta}L(\theta^*,g_0)[\theta-\theta^*,G-G_0] = 0 ~~~\forall \theta \in \hat{\Theta},G\in\hat{\mathcal{G}}, 
\end{equation} 

\vspace{0.3cm}
\noindent \textbf{Notation.} For a subset $\mathcal{X}$ of a vector space, conv($\mathcal{X}$) will denote the convex hull. For an element $x\in \mathcal{X}$, we define the star hull as in (\ref{starhull}) and adopt the shorthand star($\mathcal{X}$) $:=$ star($\mathcal{X}$,0).
\begin{equation} \label{starhull}
    \text{star}(\mathcal{X},x) = \left\{t\cdot x+(1-t)\cdot x'|x'\in \mathcal{X},t\in [0,1]\right\}
\end{equation}

\vspace{0.3cm}
\noindent \textbf{Assumption 2} (First order optimality). \\
The minimiser for the population risk satisfies the first-order optimality condition:
\begin{equation} \label{A2}
   D_{\theta}L(\theta^*,g_0)[\theta - \theta^*] \geq 0 ~~~\forall\theta\in \text{star}(\Theta,\theta^*)  
\end{equation}

\vspace{0.5cm}
\noindent \textbf{Assumption 3} (Strong convexity in prediction). \\
\noindent The population risk $L$ is strongly convex with respect to the prediction: For all $\theta \in \Theta$ and $g \in \mathcal{G}$, 
\begin{equation}
    D^2_{\theta}L(\bar{\theta},g)[\theta-\theta^*, \theta-\theta^*] \geq \lambda\|\theta - \theta^*\|_2^2 - \kappa\|g-g_0\|_2^4~~~\forall \bar{\theta}\in \text{star}(\hat{\Theta},\theta^*)
\end{equation}

\vspace{0.5cm}
\noindent \textbf{Assumption 4} (Higher-order smoothness). \\
\noindent There exist constants $\beta_1,~\beta_2>0$ such that the following derivative bounds hold: 
\begin{enumerate}[label=\alph*)]
    \item Second-order smoothness with respect to target. For all $\theta \in \hat{\Theta}$ and all $\bar{\theta} \in \text{star}(\hat{\Theta},\theta^*)$: 
    \begin{equation}
       D^2_{\theta}L(\bar{\theta},g_0)[\theta-\theta^*,\theta-\theta^*]\leq \beta_1\|\theta-\theta^*\|_2^2
    \end{equation}
    \item Higher-order smoothness. For all $\theta \in \text{star}(\hat{\Theta},\theta^*),~ g\in \mathcal{G}$, and $\bar{g}\in \text{star}(\mathcal{G},g_0)$:
    \begin{equation}
       |D_g^2D_{\theta}L(\theta^*,\bar{g})[\theta-\theta^*,g-g_0,g-g_0]|\leq \beta_2\|\theta-\theta^*\|_2\cdot \|g-g_0\|_2^2 
    \end{equation}
\end{enumerate}

\vspace{0.5cm}
Proofs that demonstrate that each assumption holds for the mDR-learner can be found in Appendix \ref{App_ex_risk_proofs}. When these assumptions hold, the excess risk bound under known nuisance parameters can written in terms of estimated nuisance parameters and their model error rates. 

Theorem 1, outlined below is an special case of Theorem 1 presented by \cite{RN22}. However, in this Theorem we contextualise their work to the mDR-learner, providing an insight into the rate robust properties of this estimator:

\vspace{0.5cm}
\noindent \textbf{Theorem 1}. Suppose that there is some $\theta^* = argmin_{\theta}L(\theta,g_0)$ such that Assumptions 1 to 4 hold (See Appendix \ref{App_ex_risk_proofs}). Then the mDR-learner produces a parameter $\hat{\theta}$ such that with probability at least $1-\delta_1-\delta_2-\delta_3-\delta_4-\delta_5-\delta_6$, with positive $\delta$'s, 
\begin{allowdisplaybreaks}
\begin{align*}
    L(\hat{\theta},g_0) - L(\theta^*,g_0) \leq & \frac{2}{1-\delta_1-\delta_2-\delta_3-\delta_4 - \delta_5-\delta_6}( \text{Rate}_{\theta}(\hat{g})  +  \frac{1}{\delta_1}\beta_{2,1}\text{Rate}_{\pi}^4 + \\
     &  \frac{1}{\delta_2}\beta_{2,2}\text{Rate}_{\pi}^2\cdot\text{Rate}^2_{\mu^1} +  \frac{1}{\delta_3}\beta_{2,3}\text{Rate}^2_{\pi}\cdot\text{Rate}_{\mu^0}^2 + \\
     & \frac{1}{\delta_4}\beta_{2,4}\text{Rate}^2_{\pi}\cdot\text{Rate}^2_{G} +  \frac{1}{\delta_5}\beta_{2,5}\text{Rate}^2_{\mu^1}\cdot\text{Rate}^2_{G}  + \frac{1}{\delta_6}\beta_{2,6}\text{Rate}^4_{G}) 
\end{align*}
\end{allowdisplaybreaks}

\noindent where
\begin{allowdisplaybreaks}
\begin{align*}
    \beta_{2,1} &= \sqrt{\E\left[\left(\frac{AC}{\hat{G}\hat{\pi}^3}\{Y-\hat{\mu}^1\}-\frac{(1-A)C}{\hat{G}(1-\hat{\pi})^3}\{Y-\hat{\mu}^0\}\right)^2\right]} \\
    \beta_{2,2} &= \sqrt{\E\left[\left(\frac{AC}{\hat{G}\hat{\pi}^2}\right)^2\right]} \\
    \beta_{2,3} &= \sqrt{\E\left[\left(\frac{(1-A)C}{\hat{G}(1-\hat{\pi})^2}\right)^2\right]} \\
    \beta_{2,4} &= \sqrt{\E\left[\left(\frac{AC}{\hat{G}^2\hat{\pi}^2}\{Y-\hat{\mu}^1\} - \frac{(1-A)C}{\hat{G}^2(1-\hat{\pi})^2}\{Y-\hat{\mu}^0\}\right)^2\right]} \\
    \beta_{2,5} &= \sqrt{\E\left[\left(\frac{AC}{\hat{G}^2\hat{\pi}}\right)^2\right]} \\
    \beta_{2,6} &= \sqrt{\E\left[\left(\frac{AC}{\hat{G}^3\hat{\pi}}\{Y-\hat{\mu}^1\} - \frac{(1-A)C}{\hat{G}^3(1-\hat{\pi})}\{Y-\hat{\mu}^0\}\right)^2\right]}
\end{align*}
\end{allowdisplaybreaks}

\vspace{0.2cm}
The bound derived in Theorem 1 comprises of seven elements; six are products of nuisance estimates and their respective convergence rates, and one contains only the convergence rate of the estimates from the second stage pseudo-outcome model. With this, when the nuisance functions are known, the bound reduces to its first term, only depending on the convergence rate of the estimates from the second stage model. However, when the nuisance functions are estimated, the bound is influenced by the convergence rate of the nuisance estimates. We also observe a rate robustness property for the outcome model estimates in the mDR-learner, as when fast rates are observed for the propensity score and missingness model estimates, the outcome model estimates can converge at slower rates without increasing this upper bound. Similar observations were shown for the DR-learner \citep{RN22}, however, once again our bound shows the extension of this robustness to the missingness model. 


\section{mDR-learner excess risk bound - Proofs}  \label{App_ex_risk_proofs}

\noindent\textbf{Assumption 1 - Proof} \\
\noindent For notational ease, we write the pseudo-outcomes created by the mDR-learner, $\phi(\pi,\mu^0,\mu^1,G)$, as $\phi(g)$, where $g$ represents the set of nuisance functions $\{\pi,\mu^0,\mu^1,G\}$. We also re-write $\phi(g)$ as:
\begin{align*}
    \phi(g) = \frac{AC}{\pi G}\{Y-\mu^1\} - \frac{(1-A)C}{(1-\pi)G}\{Y-\mu^0\} + \mu^1 - \mu^0
\end{align*}

\vspace{0.3cm}
\noindent We take the directional derivative of the MSE loss function when evaluated at its infimum with respect to the target parameter $\theta$:
\begin{align*}
    D_{\theta}L(\theta^*,g_0)[\theta - \theta^*] &= \frac{\partial}{\partial t}\{\E[(\phi(g_0) - (\theta(X)\cdot t + (1-t)\cdot \theta^*(X)))^2]\}|_{t=0} \\
    &= -2\E[(\phi(g_0) - \theta^*(X))(\theta(X) - \theta^*(X))] 
\end{align*}

\noindent Showing (\ref{A1.1_miss}) for the mDR-learner:
\begin{allowdisplaybreaks}
\begin{align*}
    D_{\pi}D_{\theta}&L(\theta^*,g_0)[\theta - \theta^*,\pi - \pi_0] \\
    &=-2\frac{\partial}{\partial t}\E\left[ \left(\frac{AC}{(\pi_0 + t \cdot(\pi- \pi_0))G_0}\{Y-\mu^1_0\}- \right.\right. \\
    &~~~~~~~~~~~~~~~~\left.\left.\left.\frac{(1-A)C}{(1-\pi_0 - t \cdot(\pi - \pi_0))G_0}\{Y-\mu^0_0\} + \mu^1_0 - \mu^0_0 - \theta^*(X)\right)(\theta(X)-\theta^*(X)) \right] \right|_{t=0} \\
    &= 2\E\left[ \left(\frac{AC}{G_0\pi_0^2}\{Y-\mu^1_0\} + \frac{(1-A)C}{G_0(1-\pi_0)^2}\{Y-\mu^0_0\} \right)(\pi - \pi_0)(\theta(X)-\theta^*(X))  \right]\\
    &= 2\E\left[\left.\E\left[ \left(\frac{AC}{G_0\pi_0^2}\{Y-\mu^1_0\} + \frac{(1-A)C}{G_0(1-\pi_0)^2}\{Y-\mu^0_0\} \right)(\pi - \pi_0)(\theta(X)-\theta^*(X))  \right|Z\right]\right]\\
    &= 2\E\left[\left.\E\left[\frac{AC}{G_0\pi_0^2}\{Y-\mu^1_0\} + \frac{(1-A)C}{G_0(1-\pi_0)^2}\{Y-\mu^0_0\}  \right|Z\right](\pi - \pi_0)(\theta(X)-\theta^*(X)) \right]\\
    &= 2\E\left[\left(\frac{1}{G_0\pi_0^2}\{\E[ACY|Z]-\E[AC|Z]\mu^1_0\} + \right.\right.\\
    &~~~~~~~~~~~\left.\left.\left.\frac{1}{G_0(1-\pi_0)^2}\{\E[(1-A)CY|Z]-\E[(1-A)C|Z]\mu^0_0\} \right)\right. (\pi - \pi_0)(\theta(X)-\theta^*(X)) \right]\\
    &= 2\E\left[\left(\frac{1}{G_0\pi_0^2}\{G_0\pi_0\mu_0^1 -G_0\pi_0\mu^1_0\} + \right.\right.\\
    &~~~~~~~~~~~\left.\left.\left.\frac{1}{G_0(1-\pi_0)^2}\{\E[G_0(1-\pi_0)\mu_0^0|Z]-G_0(1-\pi_0)\mu^0_0\} \right)\right. (\pi - \pi_0)(\theta(X)-\theta^*(X)) \right] \\
    &~~~~~~~~~~~~ \text{By conditional exchangeability and consistency}\\
    &= 0
\end{align*}
\end{allowdisplaybreaks}

\vspace{0.5cm}
\noindent Showing (\ref{A1.2_miss}) for the mDR-learner:
\begin{allowdisplaybreaks}
\begin{align*}
    D_{\mu^1}D_{\theta}&L(\theta^*,g_0)[\theta - \theta^*,\mu^1 - \mu^{1}_0] \\
    &= -2\frac{\partial}{\partial t}\E\left[ \left(\frac{AC}{G_0\pi_0}\{Y-(\mu^{1}_0 + t(\mu^{1}-\mu^{1}_0))\} - \right.\right.\\
    &~~~~~~~~~~~~~~~~\left.\left.\left.\frac{(1-A)C}{G_0(1-\pi_0)}\{Y-\mu^0_0\} + \mu^1_0 + t(\mu^1 - \mu^1_0) - \mu^0_0 - \theta^*(X)\right)(\theta(X)-\theta^*(X))  \right]\right|_{t=0} \\
    &= 2\E\left[ \left(\frac{AC}{G_0\pi_0}(\mu^{1}-\mu^{1}_0) - \mu^1 + \mu^1_0\right)(\theta(X)-\theta^*(X))  \right] \\
    &= 2\E\left[ \E\left[\left.\left(\frac{AC}{G_0\pi_0}(\mu^{1}-\mu^{1}_0) - \mu^1 + \mu^1_0\right)(\theta(X)-\theta^*(X))\right|Z  \right]\right] \\
    &= 2\E\left[ \left(\frac{\E[AC|Z]}{G_0\pi_0}(\mu^{1}-\mu^{1}_0) - \mu^1 + \mu^1_0\right)  (\theta(X)-\theta^*(X))\right] \\
    &= 2\E\left[ \left(\mu^{1}-\mu^{1}_0 - \mu^1 + \mu^1_0\right)  (\theta(X)-\theta^*(X))\right] \\
    &= 0
\end{align*}
\end{allowdisplaybreaks}

\noindent (\ref{A1.3_miss}) can be shown for the mDR-learner using the same approach as (\ref{A1.2_miss}). And showing (\ref{A1.4_miss}) for the mDR-learner:
\begin{allowdisplaybreaks}
\begin{align*}
    D_{G}D_{\theta}&L(\theta^*,g_0)[\theta - \theta^*,G - G_0] \\
     = &-2\frac{\partial}{\partial t}\E\left[ \left(\frac{AC}{(G_0+t(G-G_0)\pi_0}\{Y-\mu^{1}_0\} - \right.\right.\\
    &~~~~~~~~~~~~~~ \left.\left.\left.\frac{(1-A)C}{(G_0+t(G-G_0)(1-\pi_0)}\{Y-\mu^0_0\} +\mu^1_0 - \mu^0_0 - \theta^*(X)\right)(\theta(X)-\theta^*(X))  \right]\right|_{t=0} \\
     = &2\frac{\partial}{\partial t}\E\left[ \left(\frac{AC}{(G_0+t(G-G_0)^2\pi_0}\{Y-\mu^{1}_0\} - \right.\right.\\
    &~~~~~~~~~~~ \left.\left.\left.\frac{(1-A)C}{(G_0+t(G-G_0)^2(1-\pi_0)}\{Y-\mu^0_0\} \right)(G-G_0)(\theta(X)-\theta^*(X))  \right]\right|_{t=0} \\
     = &2\E\left[ \left(\frac{AC}{G_0^2\pi_0}\{Y-\mu^{1}_0\} - \frac{(1-A)C}{G_0^2(1-\pi_0)}\{Y-\mu^0_0\} \right)(G-G_0)(\theta(X)-\theta^*(X))  \right] \\
    =& 2\E\left[\left.\E\left[ \left(\frac{AC}{G_0^2\pi_0}\{Y-\mu^1_0\} + \frac{(1-A)C}{G_0^2(1-\pi_0)}\{Y-\mu^0_0\} \right)(G - G_0)(\theta(X)-\theta^*(X))  \right|Z,A\right]\right]\\
    =& 2\E\left[\left.\E\left[ \left(\frac{AC}{G_0^2\pi_0}\{Y-\mu^1_0\} + \frac{(1-A)C}{G_0^2(1-\pi_0)}\{Y-\mu^0_0\} \right)  \right|Z,A\right](G - G_0)(\theta(X)-\theta^*(X))\right]\\
    =& 2\E\left[ \left(\frac{1}{G_0^2\pi_0}\{\E[ACY|Z,A] -\E[AC|Z,A]\mu^1_0\} \right.\right. + \\
    &~~~~~~~\left.\left.\frac{1}{G_0^2(1-\pi_0)}\{\E[(1-A)CY|Z,A] -\E[(1-A)C|Z,A]\mu^0_0\} \right) (G - G_0)(\theta(X)-\theta^*(X))\right]\\
    =& 2\E\left[ \left(\frac{1}{G_0^2\pi_0}\{G_0\mu^1_0 -G\mu^1_0\} + \frac{1}{G_0^2(1-\pi_0)}\{G_0\mu^0_0 -G\mu^0_0\} \right) (G - G_0)(\theta(X)-\theta^*(X))\right] \\
    =& 0
\end{align*}
\end{allowdisplaybreaks}

\noindent \textbf{Assumption 2 proof}
\noindent Let us consider a $\theta_0 \in \Theta$, then we get:
\begin{allowdisplaybreaks}
\begin{align*}
     D_{\theta}L(\theta^*,g_0)[\theta - \theta_0] 
    &= -2\E\left[ (\phi(g_0)-\theta_0(X))(\theta(X)-\theta_0(X))  \right] \\
    &= -2\E[\E[(\phi(g_0)-\theta_0(X))(\theta(X) - \theta_0(X))|X]] \\
    &= -2\E[\E[\phi(g_0)-\theta_0(X)|X](\theta(X) - \theta_0(X))]\\
    &= -2\E[(\E[\phi(g_0)|X]-\theta_0(X))(\theta(X) - \theta_0(X))] \\
    &= 0 
\end{align*}
\end{allowdisplaybreaks}

\noindent As: 
\begin{align*}
    \E[\varphi(g_0;\mathcal{O})|X] &= \E\left[\left.\frac{AC}{G_0\pi_0}\{Y-\mu^1_0\} - \frac{(1-A)C}{G_0(1-\pi_0)}\{Y-\mu^0_0\} + \mu^1_0 - \mu^0_0 \right|X\right] \\
    &= \E\left[\left. \mu^1_0 - \mu^0_0 \right|X\right] \\
    &= \E\left[\left. Y(1) - Y(0) \right|X\right], ~~~~ \text{By conditional exchangeability, consistency and MAR}\\
    &= \theta_0(X)
\end{align*}

\vspace{0.3cm}
\noindent \textbf{Assumption 3 proof}
\begin{align*}
    D^2_{\theta}L(\bar{\theta},g)[\theta-\theta^*,\theta-\theta^*]
    &= \left.\frac{\partial^2}{\partial t_1 \partial t_2}\E[(\phi(g)-(\bar{\theta} + t_1(\theta - \theta^*)+ t_2(\theta - \theta^*)))^2]\right|_{t_1=t_2=0} \\
    &= \left. \frac{\partial}{\partial t_2}-2\E[(\phi(g)-(\bar{\theta} + t_2(\theta - \theta^*)))(\theta-\theta^*)]\right|_{t_2=0} \\
    &=2\E[(\theta-\theta^*)^2] 
\end{align*}

\noindent Therefore, for some $\lambda>0$,
$$ 2\E[(\theta-\theta^*)^2] \geq \lambda  \|\theta-\theta^*\|^2,$$

\vspace{0.3cm}
\noindent hence Assumption 3 holds under the condition:
$$ \lambda \leq 2  $$

\vspace{0.5cm}
\noindent \textbf{Assumption 4 proof}

\noindent \textbf{Proof (a)}
\begin{align*}
    D^2_{\theta}L(\bar{\theta},g)[\theta-\theta^*,\theta-\theta^*]
    &= \left.\frac{\partial^2}{\partial t_1 \partial t_2}\E[(\phi(g)-(\bar{\theta} + t_1(\theta - \theta^*)+ t_2(\theta - \theta^*)))^2]\right|_{t_1=t_2=0} \\
    &= \left. \frac{\partial}{\partial t_2}-2\E[(\phi(g)-(\bar{\theta} + t_2(\theta - \theta^*)))(\theta-\theta^*)]\right|_{t_2=0} \\
    &=2\E[(\theta-\theta^*)^2] 
\end{align*}

\noindent Therefore, for some $\beta_1>0$,
$$ 2\E[(\theta-\theta^*)^2] \leq \beta_1 \|\theta-\theta^*\|_2^2,$$

\vspace{0.3cm}
\noindent hence Assumption 4a holds under the condition:
$$ \beta_1 \geq 2 $$

\vspace{0.3cm}
\noindent \textbf{Proof (b)}

\noindent We need to find:
\begin{align*}
    D_{g}^2D_{\theta}&L(\theta^*,\bar{g})[\theta-\theta^*,g-g_0,g-g_0] \\
    &= \E\left[\begin{pmatrix}\pi - \pi_0 & \mu^1-\mu^1_0 & \mu^0-\mu_0 & G-G_0 \end{pmatrix}\nabla^2_{gg}\nabla_{\theta}l(\theta^*,\bar{g})\begin{bmatrix}\pi-\pi_0 \\ \mu^1-\mu^1_0 \\ \mu^0-\mu^0_0 \\ G - G_0\end{bmatrix}(\theta-\theta^*)\right]
\end{align*}

\noindent For the mDR-learner, $\nabla_{gg}^2\nabla_{\theta}l(\theta^*,\bar{g})$ takes the following form:
\begin{align*}\nabla_{gg}^2\nabla_{\theta}l(\theta^*,\bar{g}) = \begin{bmatrix}\nabla_{\pi}\nabla_{\pi}\nabla_{\theta}l(\theta,\bar{g})& \nabla_{\pi}\nabla_{\mu^1}\nabla_{\theta}l(\theta,\bar{g})& \nabla_{\pi}\nabla_{\mu^0}\nabla_{\theta}l(\theta,\bar{g}) & \nabla_{\pi}\nabla_{G}\nabla_{\theta}l(\theta,\bar{g})\\ \nabla_{\mu^1}\nabla_{\pi}\nabla_{\theta}l(\theta,\bar{g})& \nabla_{\mu^1}\nabla_{\mu^1}\nabla_{\theta}l(\theta,\bar{g})& \nabla_{\mu^1}\nabla_{\mu^0}\nabla_{\theta}l(\theta,\bar{g})& \nabla_{\mu^1}\nabla_{G}\nabla_{\theta}l(\theta,\bar{g})\\ \nabla_{\mu^0}\nabla_{\pi}\nabla_{\theta}l(\theta,\bar{g})& \nabla_{\mu^0}\nabla_{\mu^1}\nabla_{\theta}l(\theta,\bar{g})& \nabla_{\mu^0}\nabla_{\mu^0}\nabla_{\theta}l(\theta,\bar{g})& \nabla_{\mu^0}\nabla_{G}\nabla_{\theta}l(\theta,\bar{g})\\
    \nabla_{G}\nabla_{\pi}\nabla_{\theta}l(\theta,\bar{g})& \nabla_{G}\nabla_{\mu^1}\nabla_{\theta}l(\theta,\bar{g})& \nabla_{G}\nabla_{\mu^0}\nabla_{\theta}l(\theta,\bar{g})& \nabla_{G}\nabla_{G}\nabla_{\theta}l(\theta,\bar{g})
    \end{bmatrix}
\end{align*}

\vspace{0.3cm}
\noindent When substituting in the respective derivatives we get:
\begin{align*}
    &\nabla_{gg}^2\nabla_{\theta}l(\theta^*,g) = \\
    &\begin{bmatrix}
    -4\left(\frac{ac}{G\pi^3}\{y-\mu^1\} - \frac{(1-a)c}{G(1-\pi)^3}\{y-\mu^0\}\right) &  
    -\frac{2ac}{G\pi^2} & 
    -\frac{2(1-a)c}{G(1-\pi)^2} & 
    -2\left(\frac{ac}{G^2\pi^2}\{y-\mu^1\} - \frac{(1-a)c}{G^2(1-\pi)^2}\{y-\mu^0\}\right) \\  
    -\frac{2ac}{G\pi^2} & 
    0 &  
    0 & 
    -\frac{2ac}{G^2\pi} \\ 
    -\frac{2(1-a)c}{G(1-\pi)^2} & 
    0 &  
    0 & 
    \frac{2(1-a)c}{G^2\pi} \\ 
    -2\left(\frac{ac}{G^2\pi^2}\{y-\mu^1\} - \frac{(1-a)c}{G^2(1-\pi)^2}\{y-\mu^0\}\right) &  
    -\frac{2ac}{G^2\pi}  &  
    \frac{2(1-a)c}{G^2\pi} & 
    -4\left(\frac{ac}{G^3\pi}\{y-\mu^1\} - \frac{(1-a)c}{G^3(1-\pi)}\{y-\mu^0\}\right) \end{bmatrix}
\end{align*}

\vspace{0.3cm}
\noindent And therefore,
\begin{allowdisplaybreaks}
\begin{align*}
    D_{g}^2D_{\theta}&L(\theta^*,\bar{g})[\theta-\theta^*,g-g_0,g-g_0] \\  = 
    & \E\left[\{-4(\pi-\pi_0)^2\left(\frac{ac}{\bar{G}\bar{\pi}^3}\{y-\bar{\mu}^1\}-\frac{(1-a)c}{\bar{G}(1-\bar{\pi})^3}\{y-\bar{\mu}^0\}\right) - \frac{4ac}{\bar{G}\bar{\pi}^2}(\pi-\pi_0)(\mu^1-\mu^1_0)\right. -\\
    &~~~~~~ \frac{4(1-a)c}{\bar{G}(1-\bar{\pi})^2}(\pi-\pi_0)(\mu^0-\mu^0_0) - 4\left(\frac{ac}{\bar{G}^2\bar{\pi}^2}\{y-\bar{\mu}^1\} - \frac{(1-a)c}{\bar{G}^2(1-\bar{\pi})^2}\{y-\bar{\mu}^0\}\right)(\pi-\pi_0)(G-G_0) - \\
    &~~~~~~\frac{4ac}{\bar{G}^2\bar{\pi}}(\mu^1-\mu^1_0)(G-G_0) + \frac{4(1-a)c}{\bar{G}^2(1-\bar{\pi})}(\mu^0-\mu^0_0)(G-G_0) - \\
    & ~~~~~~\left. 4\left(\frac{ac}{\bar{G}^3\bar{\pi}}\{y-\bar{\mu}^1\} - \frac{(1-a)c}{\bar{G}^3(1-\bar{\pi})}\{y-\bar{\mu}^0\}\right)(G-G_0)^2 \}(\theta-\theta^*)\right] \\ 
    = &-4E\left[\left(\frac{AC}{\bar{G}\bar{\pi}^3}\{Y-\bar{\mu}^1\}-\frac{(1-A)C}{\bar{G}(1-\bar{\pi})^3}\{Y-\bar{\mu}^0\}\right)(\pi-\pi_0)^2(\theta-\theta^*)\right] - \\
    &4\E\left[\frac{AC}{\bar{G}\bar{\pi}^2}(\pi-\pi_0)(\mu^1-\mu^1_0)(\theta - \theta^*)\right] -4\E\left[\frac{(1-A)C}{\bar{G}(1-\bar{\pi})^2}(\pi-\pi_0)(\mu^0-\mu^0_0)(\theta - \theta^*)\right] - \\
    &4\E\left[\left(\frac{AC}{\bar{G}^2\bar{\pi}^2}\{Y-\bar{\mu}^1\} - \frac{(1-A)C}{\bar{G}^2(1-\bar{\pi})^2}\{Y-\bar{\mu}^0\}\right)(\pi-\pi_0)(G-G_0)(\theta - \theta^*)\right] -\\
    &4\E\left[\frac{AC}{\bar{G}^2\bar{\pi}}(\mu^1-\mu^1_0)(G-G_0)(\theta - \theta^*)\right] + 4\E\left[\frac{(1-A)C}{\bar{G}^2(1-\bar{\pi})}(\mu^0-\mu^0_0)(G-G_0)(\theta - \theta^*)\right]- \\
    &4\E\left[\left(\frac{AC}{\bar{G}^3\bar{\pi}}\{Y-\bar{\mu}^1\} - \frac{(1-A)C}{\bar{G}^3(1-\bar{\pi})}\{Y-\bar{\mu}^0\}\right)(G-G_0)^2(\theta-\theta^*)\right]
\end{align*}
\end{allowdisplaybreaks}

\vspace{0.3cm}
\noindent To identify the upper bound of this, we apply the Cauchy-Schwarz inequality iteratively, pulling out the nuisance and target norms, and use the norm inequality to simplifying each term such that they are written in terms of $L_2$ norm  
\begin{allowdisplaybreaks}
\begin{align*}
      | D_{g}^2D_{\theta}&L(\theta^*,\bar{g})[\theta-\theta^*,g-g_0,g-g_0] | \\  
      \leq  & 4\left(\E\left[\left(\frac{AC}{\bar{G}\bar{\pi}^3}\{Y-\bar{\mu}^1\}-\frac{(1-A)C}{\bar{G}(1-\bar{\pi})^3}\{Y-\bar{\mu}^0\}\right)^2(\pi-\pi_0)
      ^4\right]\right)^{1/2}\|\theta-\theta^*\|_2 + \\ 
     & 4\left(\E\left[\left(\frac{AC}{\bar{G}\bar{\pi}^2}\right)^2(\pi-\pi_0)^2(\mu^1-\mu^1_0)^2\right]\right)^{1/2}\|\theta-\theta^*\|_2 + \\
    & 4\left(\E\left[\left(\frac{(1-A)C}{\bar{G}(1-\bar{\pi})^2}\right)^2(\pi-\pi_0)^2(\mu^0-\mu^0_0)^2\right]\right)^{1/2}\|\theta-\theta^*\|_2 + \\ 
    & 4\left(\E\left[\left(\frac{AC}{\bar{G}^2\bar{\pi}^2}\{Y-\bar{\mu}^1\} - \frac{(1-A)C}{\bar{G}^2(1-\bar{\pi})^2}\{Y-\bar{\mu}^0\}\right)^2(\pi-\pi_0)^2(G-G_0)^2\right]\right)^{1/2}\|\theta-\theta^*\|_2  + \\
    &4\left(\E\left[\left(\frac{AC}{\bar{G}^2\bar{\pi}}\right)^2(\mu^1-\mu^1_0)^2(G-G_0)^2\right]\right)^{1/2}\|\theta-\theta^*\|_2  + \\
    & 4\left(\E\left[\left(\frac{AC}{\bar{G}^3\bar{\pi}}\{Y-\bar{\mu}^1\} - \frac{(1-A)C}{\bar{G}^3(1-\bar{\pi})}\{Y-\bar{\mu}^0\}\right)^2(G-G_0)^4\right]\right)^{1/2}\|\theta-\theta^*\|_2 \\ \\  
    \leq  & 4\left(\E\left[\left(\frac{AC}{\bar{G}\bar{\pi}^3}\{Y-\bar{\mu}^1\}-\frac{(1-A)C}{\bar{G}(1-\bar{\pi})^3}\{Y-\bar{\mu}^0\}\right)^4\right]^{1/2}\|\pi-\pi_0
      \|^4_8\right)^{1/2}\|\theta-\theta^*\|_2 + \\
    & 4\left(\E\left[\left(\frac{AC}{\bar{G}\bar{\pi}^2}(\pi-\pi_0)\right)^4\right]^{1/2}\|\mu^1-\mu^1_0\|_4^2\right)^{1/2}\|\theta-\theta^*\|_2 + \\
    & 4\left(\E\left[\left(\frac{(1-A)C}{\bar{G}(1-\bar{\pi})^2}(\pi-\pi_0)\right)^4\right]^{1/2}\|\mu^0-\mu^0_0\|_4^2\right)^{1/2}\|\theta-\theta^*\|_2 + \\ 
    & 4\left(\E\left[\left(\left(\frac{AC}{\bar{G}^2\bar{\pi}^2}\{Y-\bar{\mu}^1\} - \frac{(1-A)C}{\bar{G}^2(1-\bar{\pi})^2}\{Y-\bar{\mu}^0\}\right)(\pi-\pi_0)\right)^4\right]^{1/2}\|G-G_0\|^2_4\right)^{1/2}\|\theta-\theta^*\|_2  + \\
    &4\left(\E\left[\left(\frac{AC}{\bar{G}^2\bar{\pi}}(\mu^1-\mu^1_0)\right)^4\right]^{1/2}\|G-G_0\|^2_4\right)^{1/2}\|\theta-\theta^*\|_2  + \\
    & 4\left(\E\left[\left(\frac{AC}{\bar{G}^3\bar{\pi}}\{Y-\bar{\mu}^1\} - \frac{(1-A)C}{\bar{G}^3(1-\bar{\pi})}\{Y-\bar{\mu}^0\}\right)^4\right]^{1/2}\|G-G_0\|^4_8\right)^{1/2}\|\theta-\theta^*\|_2 
      \\ \\ 
    =  & 4\E\left[\left(\frac{AC}{\bar{G}\bar{\pi}^3}\{Y-\bar{\mu}^1\}-\frac{(1-A)C}{\bar{G}(1-\bar{\pi})^3}\{Y-\bar{\mu}^0\}\right)^4\right]^{1/4}\|\pi-\pi_0
      \|^2_8\|\theta-\theta^*\|_2 + \\
    & 4\E\left[\left(\frac{AC}{\bar{G}\bar{\pi}^2}(\pi-\pi_0)\right)^4\right]^{1/4}\|\mu^1-\mu^1_0\|_4\|\theta-\theta^*\|_2 + \\
    & 4\E\left[\left(\frac{(1-A)C}{\bar{G}(1-\bar{\pi})^2}(\pi-\pi_0)\right)^4\right]^{1/4}\|\mu^0-\mu^0_0\|_4\|\theta-\theta^*\|_2 + \\
    & 4\E\left[\left(\left(\frac{AC}{\bar{G}^2\bar{\pi}^2}\{Y-\bar{\mu}^1\} - \frac{(1-A)C}{\bar{G}^2(1-\bar{\pi})^2}\{Y-\bar{\mu}^0\}\right)(\pi-\pi_0)\right)^4\right]^{1/4}\|G-G_0\|_4\|\theta-\theta^*\|_2  +\\
    &4\E\left[\left(\frac{AC}{\bar{G}^2\bar{\pi}}(\mu^1-\mu^1_0)\right)^4\right]^{1/4}\|G-G_0\|_4\|\theta-\theta^*\|_2  +\\
    & 4\E\left[\left(\frac{AC}{\bar{G}^3\bar{\pi}}\{Y-\bar{\mu}^1\} - \frac{(1-A)C}{\bar{G}^3(1-\bar{\pi})}\{Y-\bar{\mu}^0\}\right)^4\right]^{1/4}\|G-G_0\|^2_8\|\theta-\theta^*\|_2 \\ \\ 
    \leq  & 4\E\left[\left(\frac{AC}{\bar{G}\bar{\pi}^3}\{Y-\bar{\mu}^1\}-\frac{(1-A)C}{\bar{G}(1-\bar{\pi})^3}\{Y-\bar{\mu}^0\}\right)^4\right]^{1/4}\|\pi-\pi_0
      \|^2_2\|\theta-\theta^*\| + \\ 
    & 4\E\left[\left(\frac{AC}{\bar{G}\bar{\pi}^2}(\pi-\pi_0)\right)^4\right]^{1/4}\|\mu^1-\mu^1_0\|_2\|\theta-\theta^*\|_2 + \\
    & 4\E\left[\left(\frac{(1-A)C}{\bar{G}(1-\bar{\pi})^2}(\pi-\pi_0)\right)^4\right]^{1/4}\|\mu^0-\mu^0_0\|_2\|\theta-\theta^*\|_2 + \\
    & 4\E\left[\left(\left(\frac{AC}{\bar{G}^2\bar{\pi}^2}\{Y-\bar{\mu}^1\} - \frac{(1-A)C}{\bar{G}^2(1-\bar{\pi})^2}\{Y-\bar{\mu}^0\}\right)(\pi-\pi_0)\right)^4\right]^{1/4}\|G-G_0\|_2\|\theta-\theta^*\|_2  +\\
    &4\E\left[\left(\frac{AC}{\bar{G}^2\bar{\pi}}(\mu^1-\mu^1_0)\right)^4\right]^{1/4}\|G-G_0\|_2\|\theta-\theta^*\|_2  +\\
    & 4\E\left[\left(\frac{AC}{\bar{G}^3\bar{\pi}}\{Y-\bar{\mu}^1\} - \frac{(1-A)C}{\bar{G}^3(1-\bar{\pi})}\{Y-\bar{\mu}^0\}\right)^4\right]^{1/4}\|G-G_0\|^2_2\|\theta-\theta^*\|_2\\ \\ 
    \leq  & 4\E\left[\left(\frac{AC}{\bar{G}\bar{\pi}^3}\{Y-\bar{\mu}^1\}-\frac{(1-A)C}{\bar{G}(1-\bar{\pi})^3}\{Y-\bar{\mu}^0\}\right)^4\right]^{1/4}\|\pi-\pi_0
      \|^2_2\|\theta-\theta^*\| + \\ 
    & 4\E\left[\left(\frac{AC}{\bar{G}\bar{\pi}^2}\right)^8\right]^{1/8}\|\pi-\pi_0\|_2\|\mu^1-\mu^1_0\|_2\|\theta-\theta^*\|_2 + \\
    & 4\E\left[\left(\frac{(1-A)C}{\bar{G}(1-\bar{\pi})^2}\right)^8\right]^{1/8}\|\pi-\pi_0\|_2\|\mu^0-\mu^0_0\|_2\|\theta-\theta^*\|_2 + \\
    & 4\E\left[\left(\frac{AC}{\bar{G}^2\bar{\pi}^2}\{Y-\bar{\mu}^1\} - \frac{(1-A)C}{\bar{G}^2(1-\bar{\pi})^2}\{Y-\bar{\mu}^0\}\right)^8\right]^{1/8}\|\pi-\pi_0\|_2\|G-G_0\|_2\|\theta-\theta^*\|_2  + \\
    &4\E\left[\left(\frac{AC}{\bar{G}^2\bar{\pi}}\right)^8\right]^{1/8}\|\mu^1-\mu^1_0\|_2\|G-G_0\|_2\|\theta-\theta^*\|_2  + \\
    & 4\E\left[\left(\frac{AC}{\bar{G}^3\bar{\pi}}\{Y-\bar{\mu}^1\} - \frac{(1-A)C}{\bar{G}^3(1-\bar{\pi})}\{Y-\bar{\mu}^0\}\right)^4\right]^{1/4}\|G-G_0\|^2_2\|\theta-\theta^*\|_2  \\ \\
    \leq  & 4\E\left[\left(\frac{AC}{\bar{G}\bar{\pi}^3}\{Y-\bar{\mu}^1\}-\frac{(1-A)C}{\bar{G}(1-\bar{\pi})^3}\{Y-\bar{\mu}^0\}\right)^2\right]^{1/2}\|\pi-\pi_0
      \|^2_2\|\theta-\theta^*\| + \\ 
    & 4\E\left[\left(\frac{AC}{\bar{G}\bar{\pi}^2}\right)^2\right]^{1/2}\|\pi-\pi_0\|_2\|\mu^1-\mu^1_0\|_2\|\theta-\theta^*\|_2 + \\
    & 4\E\left[\left(\frac{(1-A)C}{\bar{G}(1-\bar{\pi})^2}\right)^2\right]^{1/2}\|\pi-\pi_0\|_2\|\mu^0-\mu^0_0\|_2\|\theta-\theta^*\|_2 + \\
    & 4\E\left[\left(\frac{AC}{\bar{G}^2\bar{\pi}^2}\{Y-\bar{\mu}^1\} - \frac{(1-A)C}{\bar{G}^2(1-\bar{\pi})^2}\{Y-\bar{\mu}^0\}\right)^2\right]^{1/2}\|\pi-\pi_0\|_2\|G-G_0\|_2\|\theta-\theta^*\|_2  + \\
    &4\E\left[\left(\frac{AC}{\bar{G}^2\bar{\pi}}\right)^2\right]^{1/2}\|\mu^1-\mu^1_0\|_2\|G-G_0\|_2\|\theta-\theta^*\|_2  + \\
    & 4\E\left[\left(\frac{AC}{\bar{G}^3\bar{\pi}}\{Y-\bar{\mu}^1\} - \frac{(1-A)C}{\bar{G}^3(1-\bar{\pi})}\{Y-\bar{\mu}^0\}\right)^2\right]^{1/2}\|G-G_0\|^2_2\|\theta-\theta^*\|_2
\end{align*}
\end{allowdisplaybreaks}

\vspace{0.3cm}
We then re-write this as:
\begin{allowdisplaybreaks}
\begin{align*}
      |D_{g}^2D_{\theta}&L(\theta^*,\bar{g})[\theta-\theta^*,g-g_0,g-g_0]|   \leq    
     4 \left(\beta_{2,1}\|\pi-\pi_0\|_2^2\|\theta-\theta^*\|_2 + \right. \\
     & \beta_{2,2}\|\pi-\pi_0\|_2\|\mu^1-\mu^1_0\|_2\|\theta-\theta^*\|_2 +      \beta_{2,3}\|\pi-\pi_0\|_2\|\mu^0-\mu^0_0\|_2\|\theta-\theta^*\|_2 + \\ & \beta_{2,4}\|\pi-\pi_0\|_2\|G-G_0\|_2\|\theta-\theta^*\|_2  +
      \beta_{2,5}\|\mu^1-\mu^1_0\|_2\|G-G_0\|_2\|\theta-\theta^*\|_2  + \\ & \left. \beta_{2,6}\|G-G_0\|_2^2\|\theta-\theta^*\|_2\right)
\end{align*}
\end{allowdisplaybreaks}

\noindent Where 
\begin{allowdisplaybreaks}
\begin{align*}
    \beta_{2,1} &= \E\left[\left(\frac{AC}{\hat{G}\hat{\pi}^3}\{Y-\hat{\mu}^1\}-\frac{(1-A)C}{\hat{G}(1-\hat{\pi})^3}\{Y-\hat{\mu}^0\}\right)^2\right]^{1/2} \\
    \beta_{2,2} &= \E\left[\left(\frac{AC}{\hat{G}\hat{\pi}^2}\right)^2\right]^{1/2} \\
    \beta_{2,3} &= \E\left[\left(\frac{(1-A)C}{\hat{G}(1-\hat{\pi})^2}\right)^2\right]^{1/2} \\
    \beta_{2,4} &= \E\left[\left(\frac{AC}{\hat{G}^2\hat{\pi}^2}\{Y-\hat{\mu}^1\} - \frac{(1-A)C}{\hat{G}^2(1-\hat{\pi})^2}\{Y-\hat{\mu}^0\}\right)^2\right]^{1/2} \\
    \beta_{2,5} &= \E\left[\left(\frac{AC}{\hat{G}^2\hat{\pi}}\right)^2\right]^{1/2} \\
    \beta_{2,6} &= \E\left[\left(\frac{AC}{\hat{G}^3\hat{\pi}}\{Y-\hat{\mu}^1\} - \frac{(1-A)C}{\hat{G}^3(1-\hat{\pi})}\{Y-\hat{\mu}^0\}\right)^2\right]^{1/2}
\end{align*}
\end{allowdisplaybreaks}

\vspace{0.3cm}
\noindent Finally, by defining $\beta_2 = max\{\beta_{2,1},\beta_{2,2},\beta_{2,3},\beta_{2,4},\beta_{2,5},\beta_{2,6}\}$, we write the bound as:
\begin{allowdisplaybreaks}
\begin{align*}
      |D_{g}^2D_{\theta}&L(\theta^*,\bar{g})[\theta-\theta^*,g-g_0,g-g_0]|   \leq   
     4\beta_2\|\theta-\theta^*\|_2\cdot \|g-g_0\|_2^2
\end{align*}
\end{allowdisplaybreaks}

\vspace{1cm}
\noindent \textbf{Theorem 1 proof} \\
\noindent Using second-order Taylor expansion of the risk at $\hat{g}$, there exists $\bar{\theta}\in \text{star}(\mathcal{G},\theta^*)$ such that
\begin{align*}
    &L(\hat{\theta},\hat{g}) = L(\theta^*,\hat{g}) + D_{\theta}L(\theta^*,\hat{g})[\hat{\theta} - \theta^*] + \frac{1}{2}D_{\theta}^2L(\bar{\theta},\hat{g})[\hat{\theta}-\theta^*,\hat{\theta}-\theta^*] \\
    \implies & \frac{1}{2}D_{\theta}^2L(\bar{\theta},\hat{g})[\hat{\theta}-\theta^*,\hat{\theta}-\theta^*] = L(\hat{\theta},\hat{g}) - L(\theta^*,\hat{g}) - D_{\theta}L(\theta^*,\hat{g})[\hat{\theta} - \theta^*]
\end{align*}

\vspace{0.3cm}
\noindent Using strong convexity (Assumption 3) we get
\begin{align*}
    & D_{\theta}^2L(\bar{\theta},\hat{g})[\hat{\theta} - \theta^*,\hat{\theta} - \theta^*] \geq \lambda\|\hat{\theta}-\theta^*\|_2^2 \\
    \implies & \frac{\lambda}{2}\|\hat{\theta}-\theta^*\|_2^2 \leq L(\hat{\theta},\hat{g}) - L(\theta^*,\hat{g}) - D_{\theta}L(\theta^*,\hat{g})[\hat{\theta} - \theta^*]
\end{align*}

\vspace{0.3cm} 
\noindent And applying the assumed rate for $\hat{\theta}$ we obtain: 
\begin{align} \label{T1_proof1}
    \frac{\lambda}{2}\|\hat{\theta}-\theta^*\|_2^2 \leq \text{Rate}_{\theta}(\hat{g}) - D_{\theta}L(\theta^*,\hat{g})[\hat{\theta} - \theta^*]
\end{align}

\vspace{0.3cm}
\noindent Using Assumption 4, we are then able to apply an additional second-order Taylor expansion, which implies there exists $\bar{g}\in \text{star}(\mathcal{G},g_0)$ such that 
\begin{align*}
    -D_{\theta}L(\theta^*,\hat{g})[\hat{\theta} - \theta^*] =& -D_{\theta}L(\theta^*,g_0)[\hat{\theta} - \theta^*] -D_gD_{\theta}L(\theta^*,g_0)[\hat{\theta} - \theta^*,\hat{g}-g_0] \\ & -   \frac{1}{2}D_g^2D_{\theta}L(\theta^*,\bar{g})[\hat{\theta} - \theta^*,\hat{g}-g_0,\hat{g}-g_0]
\end{align*}

\vspace{0.3cm}
\noindent Which when used in combination with the orthogonality assumption (Assumption 1)
\begin{align*}
    D_{g}D_{\theta}L(\theta^*,g_0)[\hat{\theta}-\theta^*,\hat{g}-g_0] = 0
\end{align*}

\vspace{0.3cm}
\noindent We get
\begin{align*}
    -D_{\theta}L(\theta^*,\hat{g})[\hat{\theta} - \theta^*] =& -D_{\theta}L(\theta^*,g_0)[\hat{\theta} - \theta^*]  -   \frac{1}{2}D_g^2D_{\theta}L(\theta^*,\bar{g})[\hat{\theta} - \theta^*,\hat{g}-g_0,\hat{g}-g_0]
\end{align*}

\vspace{0.3cm}
\noindent Combining this with (\ref{T1_proof1}) we see
\begin{align*}
     \frac{\lambda}{2}\|\hat{\theta}-\theta^*\|_2^2 \leq \text{Rate}_{\theta}(\hat{g}) -D_{\theta}L(\theta^*,g_0)[\hat{\theta} - \theta^*]  -   \frac{1}{2}D_g^2D_{\theta}L(\theta^*,\bar{g})[\hat{\theta} - \theta^*,\hat{g}-g_0,\hat{g}-g_0]
\end{align*}

\vspace{0.3cm}
\noindent And substituting in the higher order smoothness calculation from Assumption 4b, we get
\begin{align*}
    -   \frac{1}{2}D_g^2D_{\theta}&L(\theta^*,\bar{g})[\hat{\theta} - \theta^*,\hat{g}-g_0,\hat{g}-g_0] \leq 2\left(\beta_{2,1}\|\hat{\pi}-\pi_0\|_2^2\|\hat{\theta}-\theta^*\| + \right. \\
     & \beta_{2,2}\|\hat{\pi}-\pi_0\|_2\|\hat{\mu}^1-\mu^1_0\|_2\|\hat{\theta}-\theta^*\|_2 +      \beta_{2,3}\|\hat{\pi}-\pi_0\|_2\|\hat{\mu}^0-\mu^0_0\|_2\|\hat{\theta}-\theta^*\|_2 + \\ & \beta_{2,4}\|\hat{\pi}-\pi_0\|_2\|\hat{G}-G_0\|_2\|\hat{\theta}-\theta^*\|_2  +
      \beta_{2,5}\|\hat{\mu}^1-\mu^1_0\|_2\|\hat{G}-G_0\|_2\|\hat{\theta}-\theta^*\|_2  + \\ & \left. \beta_{2,6}\|\hat{G}-G_0\|_2^2\|\hat{\theta}-\theta^*\|_2\right)
\end{align*}

\vspace{0.3cm}
\noindent And so 
\begin{allowdisplaybreaks}
\begin{align*}
     \frac{\lambda}{2}\|\hat{\theta}-\theta^*\|_2^2 \leq & \text{Rate}_{\theta}(\hat{g}) -D_{\theta}L(\theta^*,g_0)[\hat{\theta} - \theta^*]  +  2\left(\beta_{2,1}\|\hat{\pi}-\pi_0\|_2^2\|\hat{\theta}-\theta^*\|_2 + \right. \\
     & \beta_{2,2}\|\hat{\pi}-\pi_0\|_2\|\hat{\mu}^1-\mu^1_0\|_2\|\hat{\theta}-\theta^*\|_2 +      \beta_{2,3}\|\hat{\pi}-\pi_0\|_2\|\hat{\mu}^0-\mu^0_0\|_2\|\hat{\theta}-\theta^*\|_2 + \\ & \beta_{2,4}\|\hat{\pi}-\pi_0\|_2\|\hat{G}-G_0\|_2\|\hat{\theta}-\theta^*\|_2  +
      \beta_{2,5}\|\hat{\mu}^1-\mu^1_0\|_2\|\hat{G}-G_0\|_2\|\hat{\theta}-\theta^*\|_2  + \\ & \left. \beta_{2,6}\|\hat{G}-G_0\|_2^2\|\hat{\theta}-\theta^*\|_2\right)
\end{align*}
\end{allowdisplaybreaks}

\vspace{0.3cm}
\noindent Then using the AM-GM inequality, we observe that for any positive constants $\delta_1>0,\delta_2>0,\delta_3>0,\delta_4>0,\delta_5>0,\delta_6>0$, we have:
\begin{allowdisplaybreaks}
\begin{align*} 
     \frac{\lambda}{2}\|\hat{\theta}-\theta^*\|_2^2 \leq & \text{Rate}_{\theta}(\hat{g}) -D_{\theta}L(\theta^*,g_0)[\hat{\theta} - \theta^*]  +  
     \frac{1}{\delta_1}\beta_{2,1}\|\hat{\pi}-\pi_0\|_2^4 +  \\
     & \frac{1}{\delta_2}\beta_{2,2}\|\hat{\pi}-\pi_0\|_2^2\|\hat{\mu}^1-\mu^1_0\|_2^2 + \frac{1}{\delta_3}\beta_{2,3}\|\hat{\pi}-\pi_0\|_2^2\|\hat{\mu}^0-\mu^0_0\|_2^2 + \\ 
     & \frac{1}{\delta_4}\beta_{2,4}\|\hat{\pi}-\pi_0\|_2^2\|\hat{G}-G_0\|_2^2  +
       \frac{1}{\delta_5}\beta_{2,5}\|\hat{\mu}^1-\mu^1_0\|_2^2\|\hat{G}-G_0\|_2^2  + \\ 
     &  \frac{1}{\delta_6}\beta_{2,6}\|\hat{G}-G_0\|_2^4 + (\delta_1+\delta_2+\delta_3+\delta_4+\delta_5+\delta_6)\|\hat{\theta}-\theta^*\|_2^2
\end{align*}
\end{allowdisplaybreaks}

\vspace{0.3cm}
\noindent Therefore, for $\delta_1+\delta_2+\delta_3+\delta_4+\delta_5+\delta_6 <\frac{\lambda}{2}$ we see
\begin{allowdisplaybreaks}
\begin{align*}
    \|\hat{\theta}-\theta^*\|_2^2 \leq &\frac{1}{\frac{\lambda}{2}-\delta_1-\delta_2-\delta_3-\delta_4 - \delta_5-\delta_6}( \text{Rate}_{\theta}(\hat{g}) -D_{\theta}L(\theta^*,g_0)[\hat{\theta} - \theta^*]  +  \\  & \frac{1}{\delta_1}\beta_{2,1}\|\hat{\pi}-\pi_0\|_2^4 +  \frac{1}{\delta_2}\beta_{2,2}\|\hat{\pi}-\pi_0\|_2^2\|\hat{\mu}^1-\mu^1_0\|_2^2 + \frac{1}{\delta_3}\beta_{2,3}\|\hat{\pi}-\pi_0\|_2^2\|\hat{\mu}^0-\mu^0_0\|_2^2 + \\ 
     & \frac{1}{\delta_4}\beta_{2,4}\|\hat{\pi}-\pi_0\|_2^2\|\hat{G}-G_0\|_2^2  +
       \frac{1}{\delta_5}\beta_{2,5}\|\hat{\mu}^1-\mu^1_0\|_2^2\|\hat{G}-G_0\|_2^2  + \\ 
     &  \frac{1}{\delta_6}\beta_{2,6}\|\hat{G}-G_0\|_2^4)
\end{align*}
\end{allowdisplaybreaks}

\vspace{0.3cm}
\noindent And invoking Assumption 2 (\ref{A2}) we obtain
\begin{align*}
    \|\hat{\theta}-\theta^*\|_2^2 \leq &\frac{1}{\frac{\lambda}{2}-\delta_1-\delta_2-\delta_3-\delta_4 - \delta_5-\delta_6}( \text{Rate}_{\theta}(\hat{g})  +   \frac{1}{\delta_1}\beta_{2,1}\|\hat{\pi}-\pi_0\|_2^4 +  \\ & \frac{1}{\delta_2}\beta_{2,2}\|\hat{\pi}-\pi_0\|_2^2\|\hat{\mu}^1-\mu^1_0\|_2^2 + \frac{1}{\delta_3}\beta_{2,3}\|\hat{\pi}-\pi_0\|_2^2\|\hat{\mu}^0-\mu^0_0\|_2^2 + \\ 
     & \frac{1}{\delta_4}\beta_{2,4}\|\hat{\pi}-\pi_0\|_2^2\|\hat{G}-G_0\|_2^2  +
       \frac{1}{\delta_5}\beta_{2,5}\|\hat{\mu}^1-\mu^1_0\|_2^2\|\hat{G}-G_0\|_2^2  + \\ 
     &  \frac{1}{\delta_6}\beta_{2,6}\|\hat{G}-G_0\|_2^4)
\end{align*}

\noindent To identify a bound in the form of (\ref{ex-risk2}), we use another Taylor expansion which implies that there exists $\bar{\theta}\in\text{star}(\hat{\Theta},\theta^*)$ such that 
\begin{align*}
    L(\hat{\theta},g_0) - L(\theta^*,g_0) = D_{\theta}L(\theta^*,g_0)[\hat{\theta}-\theta^*] + \frac{1}{2}D_{\theta}^2L(\bar{\theta},g_0)[\hat{\theta}-\theta^*,\hat{\theta}-\theta^*]
\end{align*}

\vspace{0.3cm}
\noindent The using the smoothness bound from Assumption 4a we get:
\begin{allowdisplaybreaks}
\begin{align*}
    L(\hat{\theta},g_0) - L(\theta^*,g_0) \leq &  D_{\theta}L(\theta^*,g_0)[\hat{\theta}-\theta^*] + \frac{\beta_1}{2}\|\hat{\theta}-\theta^*\|_2^2 \\
     \leq & \frac{\beta_1}{\frac{\lambda}{2}-\delta_1-\delta_2-\delta_3-\delta_4 - \delta_5-\delta_6}( \text{Rate}_{\theta}(\hat{g})  +   \frac{1}{\delta_1}\beta_{2,1}\|\hat{\pi}-\pi_0\|_2^4 +  \\ & \frac{1}{\delta_2}\beta_{2,2}\|\hat{\pi}-\pi_0\|_2^2\|\hat{\mu}^1-\mu^1_0\|_2^2 + \frac{1}{\delta_3}\beta_{2,3}\|\hat{\pi}-\pi_0\|_2^2\|\hat{\mu}^0-\mu^0_0\|_2^2 + \\ 
     & \frac{1}{\delta_4}\beta_{2,4}\|\hat{\pi}-\pi_0\|_2^2\|\hat{G}-G_0\|_2^2  +
       \frac{1}{\delta_5}\beta_{2,5}\|\hat{\mu}^1-\mu^1_0\|_2^2\|\hat{G}-G_0\|_2^2  + \\ 
     &  \frac{1}{\delta_6}\beta_{2,6}\|\hat{G}-G_0\|_2^4) - \\ & \left(\frac{\beta_1}{\lambda-2(\delta_1+\delta_2+\delta_3+\delta_4+\delta_5+\delta_6)}-1\right)D_{\theta}L(\theta^*,g_0)[\hat{\theta}-\theta^*] 
\end{align*}
\end{allowdisplaybreaks}

\vspace{0.5cm}
\noindent Then using Assumption 2 and that $\left(\frac{\beta_1}{\lambda-2(\delta_1+\delta_2+\delta_3+\delta_4+\delta_5+\delta_6)}-1\right)\geq0$, we get
\begin{allowdisplaybreaks}
\begin{align*}
    L(\hat{\theta},g_0) - L(\theta^*,g_0) \leq & \frac{\beta_1}{\frac{\lambda}{2}-\delta_1-\delta_2-\delta_3-\delta_4 - \delta_5-\delta_6}( \text{Rate}_{\theta}(\hat{g})  +   \frac{1}{\delta_1}\beta_{2,1}\|\hat{\pi}-\pi_0\|_2^4 +  \\ & \frac{1}{\delta_2}\beta_{2,2}\|\hat{\pi}-\pi_0\|_2^2\|\hat{\mu}^1-\mu^1_0\|_2^2 + \frac{1}{\delta_3}\beta_{2,3}\|\hat{\pi}-\pi_0\|_2^2\|\hat{\mu}^0-\mu^0_0\|_2^2 + \\ 
     & \frac{1}{\delta_4}\beta_{2,4}\|\hat{\pi}-\pi_0\|_2^2\|\hat{G}-G_0\|_2^2  +
       \frac{1}{\delta_5}\beta_{2,5}\|\hat{\mu}^1-\mu^1_0\|_2^2\|\hat{G}-G_0\|_2^2  + \\ 
     &  \frac{1}{\delta_6}\beta_{2,6}\|\hat{G}-G_0\|_2^4) 
\end{align*}
\end{allowdisplaybreaks}

\noindent Which can then be expressed in terms of nuisance and target rates as: 
\begin{allowdisplaybreaks}
\begin{align*}
    L(\hat{\theta},g_0) - L(\theta^*,g_0) \leq & \frac{\beta_1}{\frac{\lambda}{2}-\delta_1-\delta_2-\delta_3-\delta_4 - \delta_5-\delta_6}( \text{Rate}_{\theta}(\hat{g})  +  \frac{1}{\delta_1}\beta_{2,1}\text{Rate}_{\pi}^4 + \\
     &  \frac{1}{\delta_2}\beta_{2,2}\text{Rate}_{\pi}^2\cdot\text{Rate}^2_{\mu^1} +  \frac{1}{\delta_3}\beta_{2,3}\text{Rate}^2_{\pi}\cdot\text{Rate}_{\mu^0}^2 + \\
     & \frac{1}{\delta_4}\beta_{2,4}\text{Rate}^2_{\pi}\cdot\text{Rate}^2_{G} +  \frac{1}{\delta_5}\beta_{2,5}\text{Rate}^2_{\mu^1}\cdot\text{Rate}^2_{G}  + \frac{1}{\delta_6}\beta_{2,6}\text{Rate}^4_{G}) 
\end{align*}
\end{allowdisplaybreaks}

\vspace{0.3cm}
\noindent Finally, using the known conditions on $\lambda$ and $\beta_1$ from assumptions 3 and 4a respectively, we can minimize our bound by setting $\beta_1=2$ and $\lambda=2$:  
\begin{allowdisplaybreaks}
\begin{align*}
    L(\hat{\theta},g_0) - L(\theta^*,g_0) \leq & \frac{2}{1-\delta_1-\delta_2-\delta_3-\delta_4 - \delta_5-\delta_6}( \text{Rate}_{\theta}(\hat{g})  +  \frac{1}{\delta_1}\beta_{2,1}\text{Rate}_{\pi}^4 + \\
     &  \frac{1}{\delta_2}\beta_{2,2}\text{Rate}_{\pi}^2\cdot\text{Rate}^2_{\mu^1} +  \frac{1}{\delta_3}\beta_{2,3}\text{Rate}^2_{\pi}\cdot\text{Rate}_{\mu^0}^2 + \\
     & \frac{1}{\delta_4}\beta_{2,4}\text{Rate}^2_{\pi}\cdot\text{Rate}^2_{G} +  \frac{1}{\delta_5}\beta_{2,5}\text{Rate}^2_{\mu^1}\cdot\text{Rate}^2_{G}  + \frac{1}{\delta_6}\beta_{2,6}\text{Rate}^4_{G}) 
\end{align*}
\end{allowdisplaybreaks}

\newpage
\section{Half-sample bootstrap confidence intervals} \label{APP_bootstrap}

Obtaining confidence intervals (CIs) for CATE estimates which are generated using non-parametric estimation techniques is challenging, with the required theoretical convergence guarantees typically not met in these settings. However, recent work by \cite{RN62} presents a solution which can estimate CIs when the CATE is both estimated using a kernel based pseudo-outcome regression and the pseudo-outcomes used in that regression take the following form:
\begin{align*}
  \mu^1(Z) - \mu^0(Z) + \beta(A,Z)(Y-\mu^A(Z)).  
\end{align*}

In the context of our work, the pseudo-outcomes from the DR-learner take this from when $\beta(A,Z) = \frac{\left(A-\pi(Z)\right)}{\pi(Z)\left(1-\pi(Z)\right)}$ and the pseudo-outcomes from the mDR-learner take this form when $\beta(A,Z) = \frac{\left(A-\pi(Z)\right)C}{\pi(Z)\left(1-\pi(Z)\right)G(A,Z)}$. The half-sample bootstrap approach then estimates CIs by approximating the distribution of  
$$R(X) = \hat{\theta}(X) - \theta(X),$$
where $\hat{\theta}(X)$ is the CATE estimate obtained using the full sample, and $\theta(X)$ is the true CATE. It does so using the conditional distribution of
$$R^*(X) = \hat{\theta}_h(X) - \hat{\theta}(X)$$
where $\hat{\theta}_h(X)$ is the CATE estimate obtained using a half sample of the data. \cite{RN62} then shows that these $R^*$'s can then be used to estimate $\hat{\lambda}^2(X)$, the variance of $\sqrt{n}R^*(X)$, and $\hat{cv}(\alpha)$, which denotes the $1-\alpha$ quantile of the distribution of the studentised statistic $\hat{S}^*(X)=\sqrt{n}\left\|\hat{\Lambda}^{-1/2}R^*(X)\right\|_{\infty}$, where $\hat{\Lambda}$ denotes the diagonal matrix with elements $\hat{\lambda}^2$. By doing so, CIs can then be constructed as:
\begin{align*}
    \hat{C}(X) = \hat{\theta}(X) \pm n^{-1/2}\hat{\lambda}(X)\hat{cv}(\alpha)
\end{align*}

This approach does not require the nuisance models to be re-run, however, for each bootstrap half-sample, the DR-learner's/mDR-learner's second stage pseudo-outcome regression must be re-run to obtain half sample estimates. We do this using a random forest, inline with the the approach used in the original work. Additionally, we implement this approach for our variations of the EP-learner, noting that the targeting step must also be re-run for each half sample bootstrap. We evaluate the performance of this approach for each of our estimators by implementing it within our simulation study. We run 500 bootstrap samples to generate the CIs and review the distribution of conditional coverages obtained within the test data. Results from this analysis can be found in Appendix \ref{Bootstrap_results}, along with a short summary of these findings. 

In addition, we implement this approach on the GBSG2 data example. However, we note two key limitations of using this technique in practice. Firstly, the implementation of this approach was only tested by \cite{RN62} using untuned random forests. This means the point estimates obtained when using this approach are likely to be worse than the point estimates obtained by the \textit{SuperLearner} approach presented in Section \ref{GBSG2} in the main text. Equally, CIs cannot be provided for the estimates obtained using the median estimator approach, as this theory does not extend to estimators constructed in this way. For these reasons, we report the findings of this additional GBSG2 analysis in Appendix \ref{APP_GB_RF} and only provide CIs for estimates obtained from one iteration of each learner.

\newpage
\section{Simulation study DGP specifications} \label{App_sim_specs}

The data in the simulation study was produced using the following data structure:
\begin{gather*}
    Z \sim Unif(-1,1)^{6}, ~~~A|Z \sim \text{Bernoulli}(\pi(Z)),~~~
    C|A,Z \sim \text{Bernoulli}(G(A,Z)) \\
    Y(0)|Z \sim \mu_0(Z) + \epsilon,~~~ Y(1)|Z,Y(0) \sim Y(0) + \theta(Z),~~~ \epsilon|Z \sim \mathcal{N}(0,1) \\
    Y = AY(0) + (1-A)Y(1)
\end{gather*}

Here, the treatment variable was generated using a propensity score $\pi(Z)$; the observed outcome was created using a combination of the unexposed outcome function $\mu_0(Z)$, the treatment effect $\theta(Z)$ and a noise parameter $\epsilon$; and the probability of this outcome being non-missing was generated using $G(A,Z)$. We also take $Z$ to be equivalent to $X$ and assume that the covariates which impact censoring occur prior to the exposure, allowing for all covariates to be adjusted for in all models. 

In each DGP, 6 covariates were created, each drawn from a uniform distribution ranging from -1 to 1. The propensity score, $\pi(Z)$, used within each DGP is defined as $\pi(Z)=expit(\sum_{i=1}^3(Z_i^2 - sin(3Z_i))/1.5)$, and each DGP uses a combination of the following censoring, unexposed outcome and CATE functions. The complex unexposed outcome function was defined as $\mu^0(Z)=\sum_{i=3}^4(Z_i/2 + 4\mathbbm{1}\{Z_i>0.5\} + sin(4Z_i))/2$, while the simple unexposed outcome function was defined as $\mu^0(Z)=Z_3 + Z_4$. The complex CATE was defined as $\theta(Z)=\mathbbm{1}\{|Z_3|>0.5\} + \mathbbm{1}\{|Z_4|>0.5\}$, while the simple CATE was defined as $\theta(Z)=0$. Then two censoring functions $G(A,Z)=expit(3 - 2.75A\sum_{i=4}^5\mathbbm{1}(|Z_i|>0.5))$ and $G(Z)=expit(3 - 2.75\sum_{\{4,5\}}\mathbbm{1}(|Z_i|>0.5))$ were defined, with the first imposing strong censoring only on the exposed arm, and the second imposing strong censoring evenly across both arms. 

The combination of functions for each DGP can been seen in the table below:

    \begin{table}[h]
    \centering
    \caption {Complexity of DGP functions}
\begin{tabular}{ccccc}
\hline
\multicolumn{1}{c}{DGP} & $\pi$ & $G$ & $\mu^0$ & $\tau$ \\ \hline
1                       &  Complex     & Complex (Depends on A)    &  Simple       &   Complex     \\
2                       &  Complex     & Complex (Does not depend on A)    & Simple        & Complex       \\
3                       &  Complex     & Complex (Depends on A)    & Complex        &  Simple      \\ \hline
\end{tabular}
\end{table}

\section{Super Learner algorithm inclusion} \label{App_sim_SL}
\begin{table}[ht]
\caption{Simulation study - Super learner algorithm libraries}
\begin{center}
\begin{tabular}{ll}
\hline
\textbf{Algorithm}   & \textbf{Tuning parameters}     \\ \hline
Mean (SL.mean)     & .     \\
& \\
Linear model (SL.lm)     & .    \\
& \\
LASSO/Elastic net (SL.glmnet) & nlambda = (100,250)\\ 
 & alpha = (0.5,1)       \\
 & \\
Random forest (SL.ranger)     & mtry = (1,2)\\ 
& min.node.size = (10,20,50)   \\
 & \\
Neural nets (SL.nnet)   & size = (1,2,5)  \\
 & \\
SVM (SL.svm)    & nu  = 1\\ 
& type = C-classification \\
& \\
Kernel KNN (SL.kernelKnn)   & h = (0.05,0.25)     \\      \hline                          
\end{tabular}
\end{center}
\end{table}

\begin{table}[ht]
\caption{GBSG2 analysis - Super learner algorithm libraries}
\begin{center}
\begin{tabular}{ll}
\hline
\textbf{Algorithm}   & \textbf{Tuning parameters}     \\ \hline
Mean (SL.mean)     & .     \\
& \\
Linear model/Generalised linear model (SL.lm/SL.glm)     & .    \\
& \\
LASSO/Elastic net (SL.glmnet) & nlambda = (100,250)\\ 
 & alpha = (0.5,1)       \\
 & \\
Random forest (SL.ranger)     & mtry = (1,2)\\ 
& min.node.size = (10,20,50)   \\
 & \\
Neural nets (SL.nnet)   & size = (1,2,5)  \\
 & \\
SVM (SL.svm)    & nu  = 1\\ 
& type = C-classification \\
& \\
Kernel KNN (SL.kernelKnn)   & h = (0.05,0.25)     \\      \hline                         \multicolumn{2}{l}{\small Neural nets, SVMs and Kernel KNN was only used in pseudo-outcome regressions}
\end{tabular}
\end{center}
\end{table}

\newpage
\section{Root mean square median error (RMSME)} \label{App_sim_RMSME}

As the DR-learner and mDR-learner use IPTWs/IPCWs, extreme propensity score/censoring probability estimates can lead to unstable CATE estimates for certain individuals. Within the simulation study, this only occurs within a small subset of the simulations and consequently, the distribution of RMSE estimates ends up skewed. For this reason, a metric such as the mean RMSE can poorly represent the performance of these estimators. 

Alternatively, we compare the estimators within the simulation study using the root mean square median error (RMSME). We calculate this metric by identifying the error in individual $j$'s CATE estimate for simulation $i$, $\epsilon_{ij}$. We then take the median of these errors across the simulations, $\epsilon_{j}^{med} = Median(\epsilon_{i,j})$. By doing so, extreme CATE estimates for each individual will become less influential and we can then estimate the root mean squared error (RMSE) using the median errors. This metric has a similar principle to the median CATE implementation option that was outlined in Section \ref{implementation}, reducing the estimators sensitivity to extreme CATE estimates and providing a metric that ignores outlier estimates.

$$RMSME = \sqrt{\frac{1}{n}\sum_{j=1}^{n}(\epsilon_{j}^{med})^2}$$

\newpage
\section{Additional simulation study results} \label{App_sim}

\begin{center}
\captionsetup{type=table}  
\caption{RMSME for each learner by sample size -  Examples 1-3} 
\begin{sideways}
\begin{tabular}{ccccccccccccccccc}
\hline
\multicolumn{1}{c}{DGP} & N & \multicolumn{1}{c}{} & \multicolumn{1}{c}{mDR}                                                & \multicolumn{1}{c}{} & \multicolumn{1}{c}{mEP}                                            & \multicolumn{1}{c}{} & \multicolumn{2}{c}{DR}                                            & \multicolumn{1}{c}{} & \multicolumn{2}{c}{EP} & \multicolumn{1}{c}{} & \multicolumn{2}{c}{T} & \multicolumn{1}{c}{} & \multicolumn{1}{c}{IPTW-IPCW} \\
\multicolumn{1}{c}{}        &             & \multicolumn{1}{c}{} & \multicolumn{1}{c}{} & \multicolumn{1}{c}{} & \multicolumn{1}{c}{} & \multicolumn{1}{c}{} & \multicolumn{1}{c}{AC} & \multicolumn{1}{c}{Imputation} & \multicolumn{1}{c}{} & \multicolumn{1}{c}{AC} & \multicolumn{1}{c}{Imputation} & \multicolumn{1}{c}{} & \multicolumn{1}{c}{AC} & \multicolumn{1}{c}{Imputation} & \multicolumn{1}{c}{} & \multicolumn{1}{c}{}   \\\hline
1       & 400   &   &  0.480 &  & 0.417  &  & 0.497 & 0.537  &  & 0.436 & 0.507 &  & \textbf{0.385} & 0.477  &  & 0.529  \\
        & 800   &   &  0.360 &  & \textbf{0.314}  &  & 0.376 & 0.410  &  & 0.329 & 0.379 &  & 0.318 & 0.376  &  & 0.486 \\
        & 1600  &   &  0.258 &  & \textbf{0.236}  &  & 0.273 & 0.305  &  & 0.249 & 0.289 &  & 0.258 & 0.307  &  & 0.493 \\
        & 3200  &   &  0.169 &  & \textbf{0.164}  &  & 0.188 & 0.219  &  & 0.178 & 0.207 &  & 0.207 & 0.250  &  & 0.557 \\
        &       &   &        &  &       &  &       &       &  &                &       &  &       &       &       &  \\
2       & 400   &   &  0.471 &  & 0.425 &  & 0.473 & 0.555 &  & 0.429 & 0.593  &  & \textbf{0.379} & 0.545 &  & 0.780 \\
        & 800   &   &  0.354 &  & 0.325 &  & 0.347 & 0.422 &  & 0.325 & 0.449  &  & \textbf{0.317} & 0.428 &  & 0.890 \\
        & 1600  &   &  0.247 &  & \textbf{0.242} &  & 0.246 & 0.309 &  & 0.244 & 0.335  &  & 0.255 & 0.340 &  & 1.129 \\
        & 3200  &   &  \textbf{0.162} &  & 0.182 &  & 0.169 & 0.222 &  & 0.185 & 0.244  &  & 0.206 & 0.280 &  & 1.562 \\
        &       &   &        &  &       &  &       &       &  &                &       &  &       &       &       &  \\
3       & 400   &   & 0.317  &  & 0.329 &  & 0.406 & 0.313 &  & 0.416 & \textbf{0.274} &  & 0.486 & 0.303 &  & 2.457 \\
        & 800   &   & 0.236  &  & 0.265 &  & 0.331 & 0.261 &  & 0.360 & \textbf{0.236} &  & 0.417 & 0.264 &  & 2.360 \\
        & 1600  &   & \textbf{0.197}  &  & 0.236 &  & 0.289 & 0.241 &  & 0.325 & 0.231 &  & 0.346 & 0.254 &  & 2.506 \\
        & 3200  &   & \textbf{0.158}  &  & 0.206 &  & 0.261 & 0.219 &  & 0.293 & 0.231 &  & 0.300 & 0.215 &  & 2.902 \\
 \hline
 \multicolumn{17}{l}{\small\textit{DGP = Data generating process; N = Training sample size; AC = Available case; RMSME = Root mean squared median error.}}
\end{tabular}
\end{sideways}
\end{center}

\newpage
\begin{figure}[H]
\begin{center}
\includegraphics[scale=0.55]{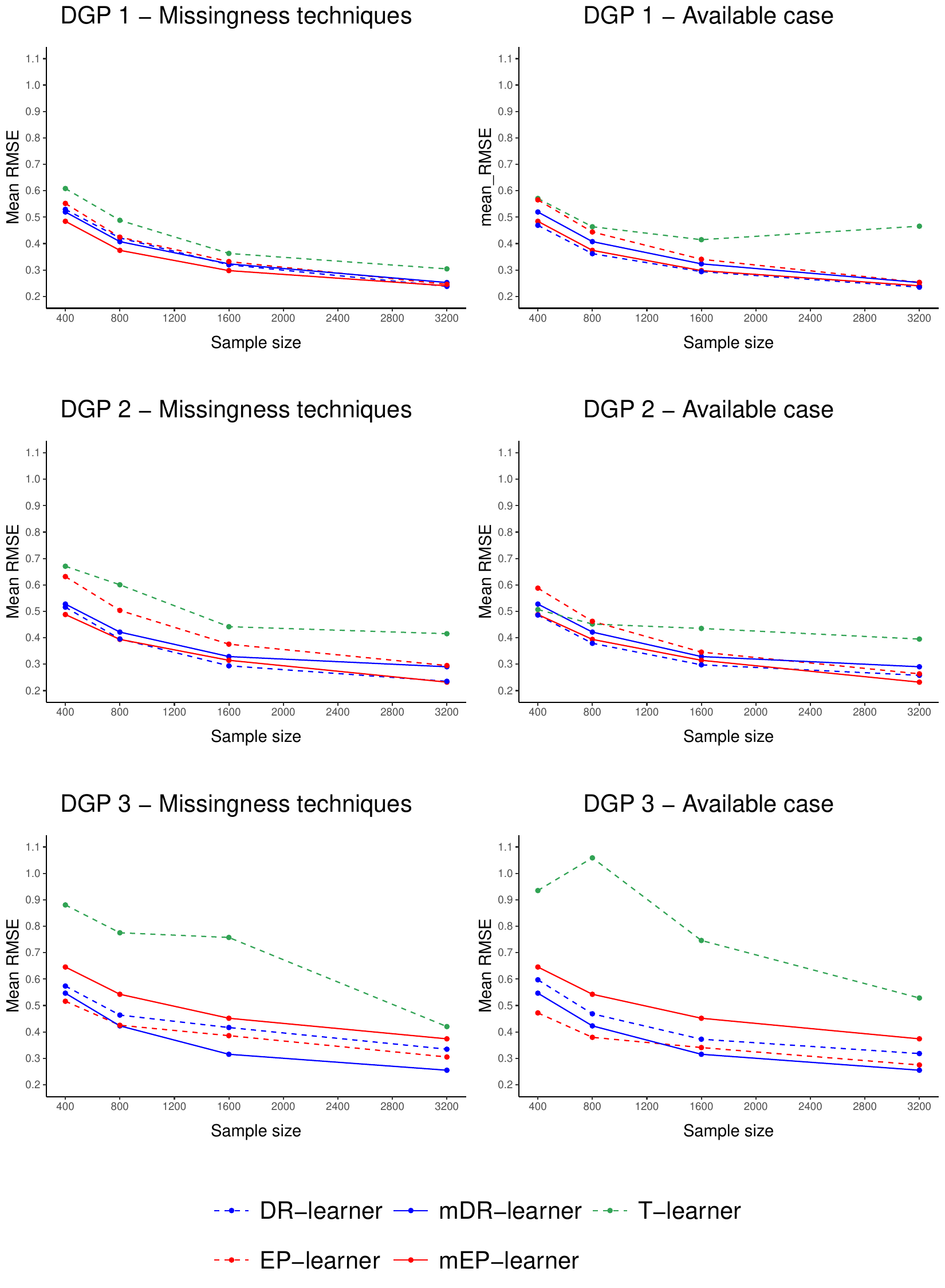} 
\end{center}
\caption{Mean RMSE for mDR-learner, mEP-learner, DR-learner, EP-learner and T-learner in three DGPs plotted by training sample size. Plots in the left column compare the mDR-learner and mEP-learner to the DR-learner, EP-learner and T-learner when used in combination with an outcome imputation model in DGP 1, 2 and 3 respectively. Plots in the right column compare the mDR-learner and mEP-learner to the available case versions of the DR-learner, EP-learner and T-learner in DGP 1, 2 and 3 respectively.}
\label{Sims_example_1_new2}
\end{figure}

\newpage
\begin{center}
\captionsetup{type=table}  
\caption{Mean RMSE for each learner by sample size -  Examples 1-3}  
\begin{sideways}
\begin{tabular}{ccccccccccccccccc}
\hline
\multicolumn{1}{c}{DGP} & N & \multicolumn{1}{c}{} & \multicolumn{1}{c}{mDR}                                                & \multicolumn{1}{c}{} & \multicolumn{1}{c}{mEP}                                            & \multicolumn{1}{c}{} & \multicolumn{2}{c}{DR}                                            & \multicolumn{1}{c}{} & \multicolumn{2}{c}{EP} & \multicolumn{1}{c}{} & \multicolumn{2}{c}{T} & \multicolumn{1}{c}{} & \multicolumn{1}{c}{IPTW-IPCW} \\
\multicolumn{1}{c}{}        &             & \multicolumn{1}{c}{} & \multicolumn{1}{c}{} & \multicolumn{1}{c}{} & \multicolumn{1}{c}{} & \multicolumn{1}{c}{} & \multicolumn{1}{c}{AC} & \multicolumn{1}{c}{Imputation} & \multicolumn{1}{c}{} & \multicolumn{1}{c}{AC} & \multicolumn{1}{c}{Imputation} & \multicolumn{1}{c}{} & \multicolumn{1}{c}{AC} & \multicolumn{1}{c}{Imputation} & \multicolumn{1}{c}{} & \multicolumn{1}{c}{}   \\\hline
1       & 400   &   & 0.520 &  & 0.484 &  & \textbf{0.469} & 0.529 &  & 0.565 & 0.552  &  & 0.571 & 0.608 &  & 0.659  \\
        & 800   &   & 0.407 &  & 0.374 &  & \textbf{0.362} & 0.421 &  & 0.444 & 0.424  &  & 0.464 & 0.488 & & 0.606 \\
        & 1600  &   & 0.323 &  & 0.298 &  & \textbf{0.294} & 0.320 &  & 0.341 & 0.332  &  & 0.415 & 0.363 & & 0.609 \\
        & 3200  &   & 0.253 &  & 0.240 &  & \textbf{0.235} & 0.238 &  & 0.253 & 0.246  &  & 0.466 & 0.304 & & 0.656 \\
        &       &   &       &  &       &  &       &       &  &                &       &  &       &       &       &  \\
2       & 400   &   & 0.528 &  & 0.488 &  & \textbf{0.486} & 0.516 &  & 0.588 & 0.631  &  & 0.507 & 0.671 & & 1.059 \\
        & 800   &   & 0.421 &  & 0.394 &  & \textbf{0.379} & 0.395 &  & 0.462 & 0.504  &  & 0.452 & 0.601 & & 1.150 \\
        & 1600  &   & 0.329 &  & 0.315 &  & 0.297 & \textbf{0.294} &  & 0.346 & 0.376  &  & 0.435 & 0.442 & & 1.388 \\
        & 3200  &   & 0.290 &  & \textbf{0.232} &  & 0.258 & 0.235 &  & 0.263 & 0.295  &  & 0.395 & 0.415 & & 1.867 \\
        &       &   &       &  &       &  &       &       &  &                &        &  &       &       & &  \\
3       & 400   &   & 0.547 &  & 0.646 &  & 0.597 & 0.574 &  & \textbf{0.472} & 0.516  &  & 0.935 & 0.881 & & 2.865 \\
        & 800   &   & 0.423 &  & 0.543 &  & 0.468 & 0.464 &  & \textbf{0.380} & 0.425  &  & 1.059 & 0.775 & & 2.727 \\
        & 1600  &   & \textbf{0.315} &  & 0.452 &  & 0.372 & 0.417 &  & 0.341 & 0.386  &  & 0.746 & 0.758 & & 2.845 \\
        & 3200  &   & \textbf{0.255} &  & 0.374 &  & 0.318 & 0.335 &  & 0.275 & 0.305  &  & 0.528 & 0.420 & & 3.245 \\
 \hline
  \multicolumn{17}{l}{\small\textit{DGP = Data generating process; N = Training sample size; AC = Available case; RMSE = Root mean squared error.}}
\end{tabular}
\end{sideways}
\end{center}

\newpage
\section{95\% confidence interval coverage results}\label{Bootstrap_results}

The performance of the half sample bootstrap procedure varied across our simulation settings. In settings with a complex CATE (DGP 1 and 2), the marginal coverage increased as sample size grew, while for settings with a simple CATE (DGP 3), the procedure provided good coverage in lower sample sizes, but this coverage dropped as the sample size increased. We also note that for settings with a complex CATE, the conditional coverage for some individuals was very poor, with the minimum 95\% CI coverage in single digits. Meanwhile, in settings where the CATE was simple, the minimum conditional coverage was higher, but did begin to fall as the sample size increased. Finally, we note that all variations of our two estimators performed consistently and similar trends in CI coverage were seen for all estimators.

\begin{table}[ht]
\centering
\caption{Summary of conditional 95\% confidence interval coverage for \\ each learner by sample size -  Example 1}
\begin{tabular}{cccccc}
\hline
Learner              & Sample size          & Mean                 & Median               & SD                          & Range                          \\
 &  & \% & \% &  & \multicolumn{1}{c}{(Min, Max)} \\ \hline
mDR              & 400   & 79.0  & 95.8  & 24.1  & (3.0, 100.0)    \\
                 & 800   & 91.2  & 98.8  & 15.1  & (4.4, 100.0)    \\
                 & 1600  & 95.7  & 100.0  & 10.2  & (10.6, 100.0)    \\
                 & 3200  & 98.0  & 100.0  & 7.1  & (16.8, 100.0)    \\
                 &       &   &   &   &     \\
mEP              & 400   & 87.0  & 98.4  & 20.4  & (2.0, 100.0)    \\
                 & 800   & 90.2  & 99.8  & 18.4  & (1.8, 100.0)    \\
                 & 1600  & 93.2  & 99.8  & 14.5  & (3.2, 100.0)    \\
                 & 3200  & 96.1  & 99.8  & 9.2   & (6.8, 100.0)    \\
                 &       &   &   &   &     \\
DR - AC          & 400   & 69.7  & 87.2  & 31.8  & (1.2, 100.0)    \\
                 & 800   & 84.1  & 98.0  & 24.6  & (1.2, 100.0)    \\
                 & 1600  & 90.4  & 99.8  & 19.2  & (3.2, 100.0)    \\
                 & 3200  & 95.7  & 100.0  & 11.7  & (5.2, 100.0)    \\
                 &       &   &   &   &     \\
DR - Imputation  & 400   & 64.2  & 75.7  & 34.2  & (0.4, 100.0)    \\
                 & 800   & 80.1  & 95.8  & 28.5  & (1.0, 100.0)    \\
                 & 1600  & 87.2  & 99.6  & 23.7  & (0.8, 100.0)    \\
                 & 3200  & 92.7  & 100.0  & 16.8  & (3.0, 100.0)   \\
                 &       &   &   &   &     \\
EP - AC          & 400   & 84.4  & 97.2  & 23.6  & (2.6, 100.0)    \\
                 & 800   & 88.5  & 99.8  & 21.4  & (2.4, 100.0)    \\
                 & 1600  & 91.5  & 99.6  & 17.4  & (2.2, 100.0)    \\
                 & 3200  & 94.2  & 99.8  & 12.2  & (4.4, 100.0)    \\
                 &       &   &   &   &     \\
EP - Imputation  & 400   & 80.4  & 94.0  & 26.5  & (2.0, 100.0)    \\
                 & 800   & 87.4  & 99.4  & 22.5  & (4.0, 100.0)    \\
                 & 1600  & 91.4  & 99.6  & 17.7  & (2.6, 100.0)    \\
                 & 3200  & 95.0  & 99.8  & 12.3  & (5.2, 100.0)    \\ \hline
\multicolumn{6}{l}{\small\textit{AC = Available case; SD = Standard deviation across conditional coverage percentages}}
\end{tabular}
\end{table}

\newpage
\begin{table}[ht]
\centering
\caption{Summary of conditional 95\% confidence interval coverage for \\ each learner by sample size -  Example 2}
\begin{tabular}{cccccc}
\hline
Learner              & Sample size          & Mean                 & Median               & SD                          & Range                          \\
 &  & \% & \% &  & \multicolumn{1}{c}{(Min, Max)} \\ \hline
mDR              & 400   & 83.5  & 97.0  & 19.7  & (7.4, 100.0)    \\
                 & 800   & 92.5  & 98.8  & 12.6  & (11.0, 100.0)    \\
                 & 1600  & 97.2  & 100.0  & 7.1  & (16.2, 100.0)    \\
                 & 3200  & 98.9  & 100.0  & 4.6  & (25.2, 100.0)    \\
                 &       &   &   &   &     \\
mEP              & 400   & 83.0  & 96.4  & 23.6  & (3.4, 100.0)    \\
                 & 800   & 86.6  & 99.4  & 22.5  & (3.4, 100.0)    \\
                 & 1600  & 88.8  & 99.0  & 18.8  & (3.2, 100.0)    \\
                 & 3200  & 90.0  & 99.0  & 16.9  & (3.8, 100.0)    \\
                 &       &   &   &   &     \\
DR - AC          & 400   & 70.1  & 83.2  & 30.8  & (1.4, 100.0)    \\
                 & 800   & 82.8  & 98.2  & 26.0  & (2.2, 100.0)    \\
                 & 1600  & 88.5  & 99.4  & 19.9  & (3.8, 100.0)    \\
                 & 3200  & 93.1  & 99.4  & 13.0  & (4.2, 100.0)    \\
                 &       &   &   &   &     \\
DR - Imputation  & 400   & 60.9  & 71.4  & 32.6  & (0.8, 100.0)    \\
                 & 800   & 74.6  & 94.0  & 32.8  & (0.2, 100.0)    \\
                 & 1600  & 80.1  & 97.6  & 29.1  & (0.8, 100.0)    \\
                 & 3200  & 84.9  & 99.0  & 23.1  & (2.0. 100.0)    \\
                 &       &   &   &   &     \\
EP - AC          & 400   & 81.2  & 96.2  & 25.0  & (2.8, 100.0)    \\
                 & 800   & 85.8  & 99.2  & 23.6  & (2.6, 100.0)    \\
                 & 1600  & 87.8  & 98.8  & 20.4  & (3.4, 100.0)    \\
                 & 3200  & 88.1  & 98.6  & 19.8  & (3.2, 100.0)    \\
                 &       &   &   &   &     \\
EP - Imputation  & 400   & 69.3  & 85.8  & 32.2  & (1.2, 100.0)    \\
                 & 800   & 77.4  & 97.0  & 30.9  & (2.0, 100.0)    \\
                 & 1600  & 81.2  & 97.0  & 26.8  & (0.6, 100.0)    \\
                 & 3200  & 86.2  & 98.6  & 21.2  & (3.0, 100.0)    \\ \hline
\multicolumn{6}{l}{\small\textit{AC = Available case; SD = Standard deviation across conditional coverage percentages}}
\end{tabular}
\end{table}

\newpage
\begin{table}[ht]
\centering
\caption{Summary of conditional 95\% confidence interval coverage for \\ each learner by sample size -  Example 3}
\begin{tabular}{cccccc}
\hline
Learner              & Sample size          & Mean                 & Median               & SD                          & Range                          \\
 &  & \% & \% &  & \multicolumn{1}{c}{(Min, Max)} \\ \hline
mDR              & 400   & 99.5  & 99.8  & 0.8  & (93.4, 100.0)    \\
                 & 800   & 99.5  & 99.8  & 0.9  & (91.4, 100.0)    \\
                 & 1600  & 99.2  & 99.8  & 1.8  & (81.2, 100.0)    \\
                 & 3200  & 98.9  & 100.0  & 2.9  & (72.0, 100.0)    \\
                 &       &   &   &   &     \\
mEP              & 400   & 98.3  & 98.8  & 1.6  & (90.6, 100.0)    \\
                 & 800   & 98.5  & 99.4  & 2.2  & (85.6, 100.0)    \\
                 & 1600  & 97.2  & 99.4  & 4.6  & (72.6, 100.0)    \\
                 & 3200  & 95.9  & 99.8  & 8.0  & (55.8, 100.0)    \\
                 &       &   &   &   &     \\
DR - AC          & 400   & 96.9  & 98.2  & 3.8  & (77.6, 100.0)    \\
                 & 800   & 96.6  & 98.8  & 5.8  & (64.6, 100.0)    \\
                 & 1600  & 94.8  & 99.2  & 9.6  & (38.0, 100.0)    \\
                 & 3200  & 93.7  & 99.8  & 13.7  & (17.6, 100.0)    \\
                 &       &   &   &   &     \\
DR - Imputation  & 400   & 98.6  & 99.2  & 1.9  & (88.0, 100.0)    \\
                 & 800   & 98.4  & 99.6  & 3.5  & (73.4, 100.0)    \\
                 & 1600  & 96.9  & 99.8  & 7.1  & (45.4, 100.0)    \\
                 & 3200  & 95.3  & 99.8  & 11.3  & (16.8, 100.0)    \\
                 &       &   &   &   &     \\
EP - AC          & 400   & 97.5  & 98.4  & 2.9  & (84.0, 100.0)    \\
                 & 800   & 96.9  & 99.2  & 5.2  & (70.6, 100.0)    \\
                 & 1600  & 94.3  & 99.0  & 9.3  & (48.2, 100.0)    \\
                 & 3200  & 92.1  & 99.6  & 14.7  & (20.0, 100.0)    \\
                 &       &   &   &   &     \\
EP - Imputation  & 400   & 99.8  & 100.0  & 0.3  & (97.8, 100.0)    \\
                 & 800   & 99.8  & 100.0  & 0.6  & (95.4, 100.0)    \\
                 & 1600  & 99.1  & 100.0  & 2.1  & (84.6, 100.0)    \\
                 & 3200  & 98.2  & 100.0  & 4.6  & (58.4, 100.0)    \\\hline
\multicolumn{6}{l}{\small\textit{AC = Available case; SD = Standard deviation across conditional coverage percentages}}
\end{tabular}
\end{table}

\newpage
\section{GBSG2 analysis using random forests for pseudo-outcome regressions} \label{APP_GB_RF}

\begin{figure}[!htb]
\begin{center}
\includegraphics[scale=0.55]{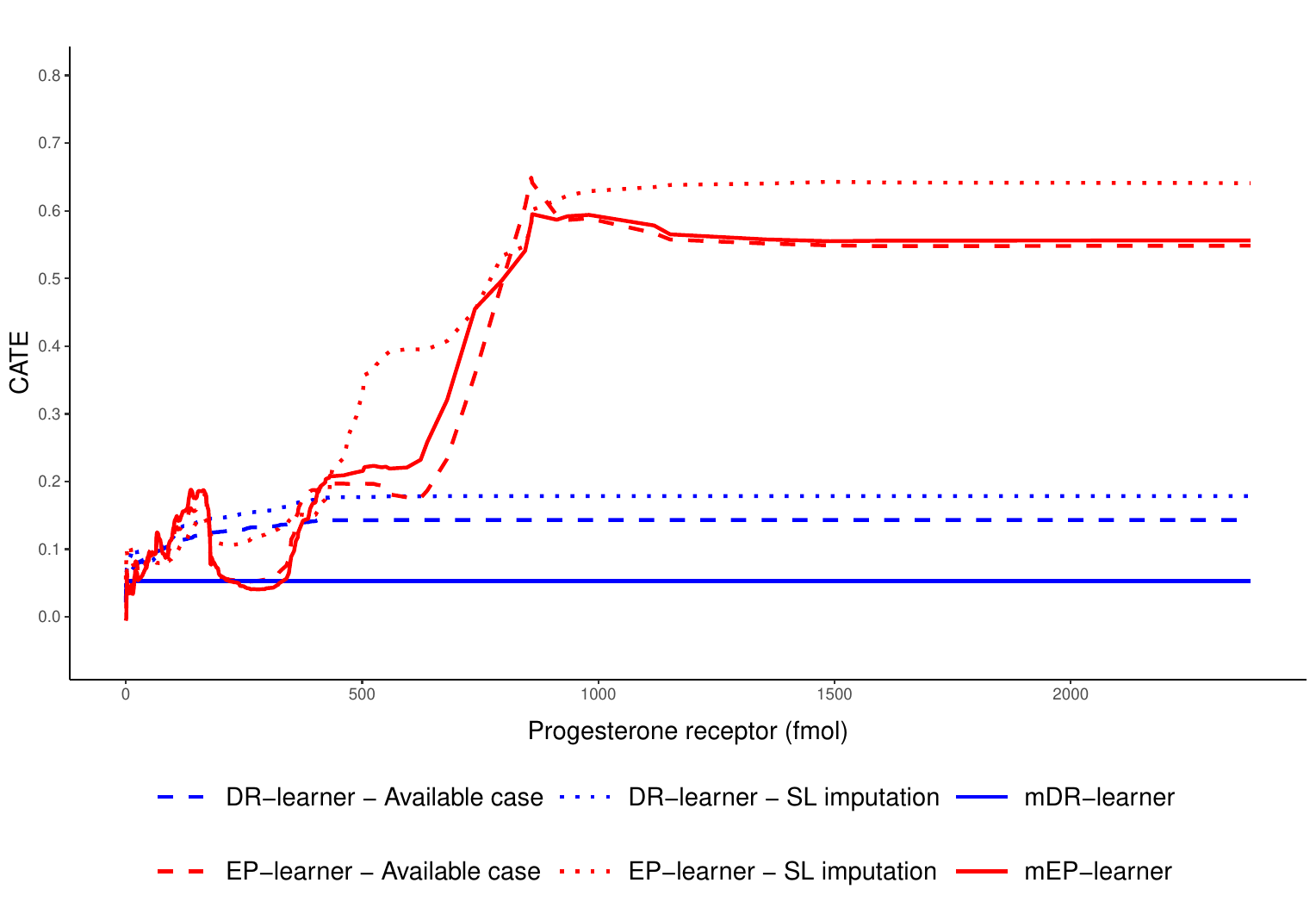} 
\end{center}
\caption{Median CATE estimates plotted by progesterone receptor (fmol/l).}
\label{Data_ex_1_rf}
\end{figure}

\newpage
\begin{figure}[!htb]
\begin{center}
\includegraphics[scale=0.45]{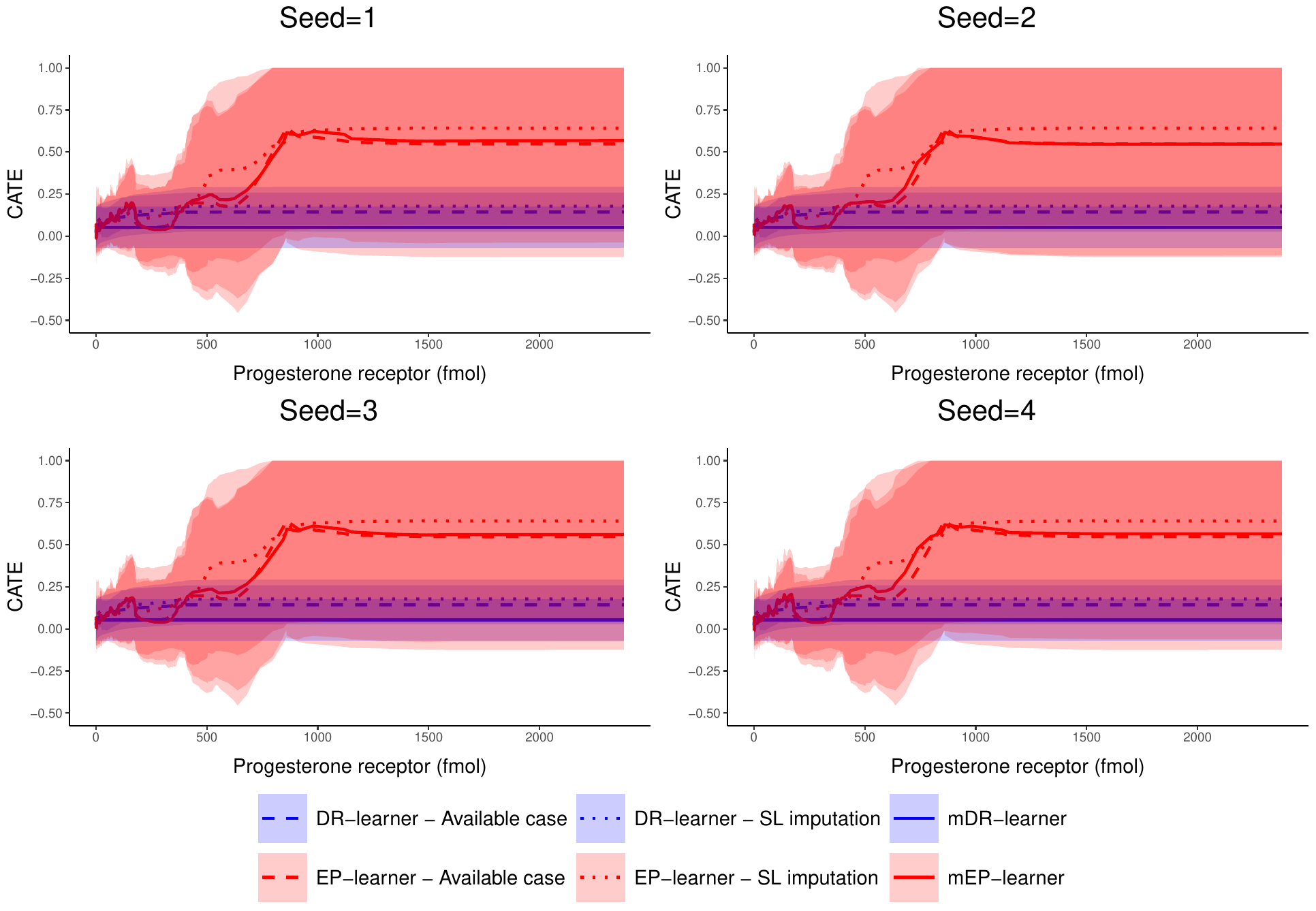} 
\end{center}
\caption{CATE estimates from single cross-fitting seeds plotted by progesterone receptor (fmol/l).}
\label{Data_ex_2_rf}
\end{figure}

\newpage
\section{mDR-learner - Longitudinal data} \label{Panel_ext}
\subsection{Setting} \label{Ex_setting}

Consider a setting with $L$ post baseline visits, a binary treatment $A$ recorded at baseline (visit $0$) and an outcome $Y$ recorded at the final visit (visit $L$). In this setting, missingness can occur at each visit $j$ for $j\in\{1,...L\}$, with the indicator for remaining non-missing at visit $j$ defined as $C_{j}$. 

Then let the covariate set $Z$ contain information which is sufficient to control for all confounders between the treatment $A$ and the outcome $Y$, and also contains sufficient information for missing outcomes to be assumed MAR. Covariate information is recorded at each visit $j$ for $j\in\{0,...,L-1\}$ and is defined as $Z_{j}$. Additionally, $C_j$ is identified prior to the recording of covariates $Z_{j}$ and the covariate and outcome missingness histories up to visit $j$ are defined as $\bar{\textbf{Z}}_{j}$ and $\bar{\textbf{C}}_{j}$ respectively, with $\bar{\textbf{Z}}_{j} = \{Z_{0},...,Z_{j}\}$ and $\bar{\textbf{C}}_{j} = \{C_{0},...,C_{j}\}$. 

We consider balanced data, with each individual following the same visit schedule, and focus on learning the CATE, $\theta(X)=\E[Y(1)|X] - \E[Y(0)|X]$, where $X$ are the baseline covariates in which heterogeneity is of interest, with $X\subseteq Z$.

\subsubsection{Identifiability}

For the CATE to be identifiable in this setting, we require A1-A5 from Section \ref{Background_miss}, along with additional outcome missingness assumptions at each visit, assuring that the missing outcome data can be considered MAR even when the relationship between the missingness indicator and outcome changes over time.

We define the MAR assumption for each time $t$, $t \in \{1,...,L\}$, as conditional independence between the outcome missingness indicator at time $t$, $C_t$, and the observed outcome $Y$ given the treatment, $A$, the baseline covariates, $Z_{0}$, and the time varying covariates up to time $t-1$; ($Y\indep C_t | A, Z_{0}, \bar{\textbf{Z}}_{t-1}$). 

By doing so, we can write the CATE in terms of iterative expectations as 
$$\E[\mu^1(L) - \mu^0(L)|X]$$

\textit{where $\mu^a(0) = y$ and $\mu^a(t) = \E[\mu^a(t-1)|A=a,\Bar{\textbf{C}}_{L+1-t}=\textbf{1},\Bar{Z}_{L-t}]$ for $t \in \{1,...,L\}$, $a\in\{0,1\}$.} \\

\subsubsection{Algorithm}
Using this statistical estimand, the panel data extension to the mDR-learner can be constructed by defining pseudo-outcomes using the EIF of the MSE for the CATE in this setting (\ref{pse_eq_L_1}):
\begin{align} \label{pse_eq_L_1}
    Y_{mDR-L} = \sum_{t=1}^L\left(\frac{(a-\hat{\pi})\prod_{k=1}^tc_{k}}{\hat{\pi}(1-\hat{\pi})\hat{G}(t)}\left\{\mu^a(L-t) - \mu^a(L+1-t)\right\}\right) + \hat{\mu}^1(L) - \hat{\mu}^0(L)
\end{align} 
\textit{where} $G(0) = 1$ and $G(j) = \prod_{i=1}^{j} P(C_i=1|A=1,\bar{\textbf{C}}_{i-1} = \textbf{1},\bar{\textbf{Z}}_{i-1})$ for $j \in (1, ... ,L)$. 

When $L=1$, these pseudo-outcomes reduce down to the mDR-learner pseudo-outcomes from 3.1. However, when $L>1$, the pseudo-outcomes incorporate post baseline measurements through time varying missingness/outcome predictions, with CATE estimates obtained by regressing the pseudo-outcomes against $X$. We present the algorithm for the mDR-learner panel data extension using a K-fold cross-fitting procedure below.

\begin{algorithm}
\caption{mDR-learner - $L$ post basline visits}\label{alg:mDR_L}
\begin{algorithmic}[1] 
\State Split the data randomly into $K$ (e.g., 10) equal sized folds of $n$ observations from $O = (A,Z_{0},C_{1}Z_{1},...,C_{L-1}Z_{L-1},C_{L}Y)$, denoted $D_{1},\ldots,D_{K}$.
\State For $j \in 1,...,K$ and using all folds $\{D_{i},i=1,\dots,K, i\neq j\}$ except $D_{j}$, train models for 
\begin{eqnarray}\label{alph}
\pi(Z)&=&P[A=1|Z], \text{(propensity score)}\\
G(t) &=& \prod_{k=1}^{t} P[C_k=1|A=a,\bar{\textbf{C}}_{k-1} = \textbf{1},\bar{\textbf{Z}}_{k-1}],~for~t \in (1, ... ,L), \\
&&\text{missingness model, with }G(0) = 1 \\
\mu^a(t) &=& \E(\mu^a(t-1)|A=a,\bar{\textbf{C}}_{L+1-t} = \textbf{1},\bar{\textbf{Z}}_{L-t}) ~\text{for}~ t \in  (1,...,L), a\in\{0,1\} \\
&& \text{(conditional outcome models in treated/untreated)} 
\end{eqnarray}
\State For all individuals in $D_{j}$ ($j \in 1,...,K$), obtain predictions $\hat{\pi}$ and $\hat{G}(t), ~\hat{\mu}^0(t)$ and $\hat{\mu}^1(t)$, $t \in \{0,...,L\}$, based on the models fitted in the remaining folds.
\State Construct the pseudo outcomes for each individual in the data using (\ref{pse_eq_L_1}).
\State Regress the pseudo outcomes $Y_{mDR-L}$ on covariates $X$, and obtain predictions of $\theta(X)$:  
$$ \hat{\theta}_{mDR-L}(X) = \hat{\E}[Y_{mDR-L}|X]$$
\end{algorithmic}
\end{algorithm}

\subsection{Implementation notes} \label{Application_ext}

Obtaining outcome/missingness predictions is more complicated in the panel data extension of the mDR-learner as they must incorporate post baseline measurements. Missingness predictions can be obtained by running a pooled logistic regression model, obtaining conditional outcome missingness probabilities (conditional on being non-missing up to that time), which in turn can be used to obtain each $\hat{G}(t)$. Meanwhile, estimating the outcome predictions $\hat{\mu}^a(t)$ for $t \in \{1,...,L\}$, $a\in\{0,1\}$ requires an iterative process, as each $\mu^a(t)$ depends on $\mu^a(t-1)$. 

When $L=1$, this translates to estimating $\mu^a(1)=\E[Y|A=a,C=1,Z]$ for $a\in\{0,1\}$ (i.e., the mDR-learner in Section \ref{DR_ext}), and is straightforward as $Y$ is observed for all individuals that remain non-missing ($C=1$). However,  for $L>1$, the outcome, $\mu^a(t-1)$, no longer represents an observed measure. 

A naive solution would be to estimate the required outcomes $\hat{\mu}^a(t-1)$ using the outcome model from the previous iteration. However, by doing so, errors from the previous model will propagate through this iterative process. Additionally, as the estimates of $\mu^a(t-1)$ are only trained using individuals who were observed up to time $N+1-t$, this population may not be representative of the population that we need to obtain outcome estimates for (i.e. people who were observed at time $N-t$). If this is not addressed, the outcome predictions for individuals who have poor representation at that time point may end up biased.

This issue has been discussed in the context of the ATE by \cite{RN52} where they use a targeted learning based estimator to account for the population shifts at each time. However, we consider this issue in the context of the CATE. In this setting we require our outcome function estimates to be conditional on a set of individual level covariates. For this reason, we adjust for the disparity in the training/target populations by running a DR-learner at each time interval. This process generates pseudo-outcomes at each interval by taking the plug-in outcome prediction and adding a weighted error to the outcome estimates, where the weight is defined using the probability of being observed up to that time. A version of this sequential DR-learner approach is implemented on \url{https://github.com/Matt-Pryce/mDR-learner_mEP-learner}.

\end{document}